\documentclass{article} 
\usepackage{times}
\usepackage{hyperref}
\usepackage{url}
\usepackage{npbayes}
\usepackage[ruled,vlined]{algorithm2e}
\usepackage[sort&compress, numbers]{natbib}
\usepackage{subfigure}
\usepackage{booktabs,dcolumn}
\usepackage{framed}

\title{Streaming Variational Bayes}

\author{
Tamara Broderick,
Nicholas Boyd,
Andre Wibisono, \\
Ashia C.~Wilson,
Michael I.~Jordan
}

\begin{document}

\maketitle

\begin{abstract}
We present SDA-Bayes, a framework for (S)treaming, (D)istributed,
(A)synchronous computation of a Bayesian posterior. The framework
makes streaming updates to the estimated posterior according to 
a user-specified approximation batch primitive.  We demonstrate 
the usefulness of our framework, with variational Bayes (VB) as 
the primitive, by fitting the latent Dirichlet allocation model 
to two large-scale document collections.  We demonstrate the advantages
of our algorithm over stochastic variational inference
(SVI) by comparing the two after a single pass through 
a known amount of data---a case where SVI may be
applied---and in the streaming setting, where SVI does not apply.
\end{abstract}

\section{Introduction}

Large, streaming data sets are increasingly the norm in
science and technology.  Simple descriptive statistics 
can often be readily computed with a constant number of 
operations for each data point in the streaming setting, without the need to revisit 
past data or have advance knowledge of future data. But
these time and memory restrictions are not generally available
for the complex, hierarchical models that practitioners 
often have in mind when they collect large data sets.  
Significant progress on scalable learning procedures has been 
made in recent years~\cite[e.g.,][]{niu:2011:hogwild,kleiner:2012:big}.
But the underlying models remain simple, and the inferential
framework is generally non-Bayesian.  The advantages of the 
Bayesian paradigm (e.g., hierarchical modeling, coherent treatment 
of uncertainty) currently seem out of reach in the Big Data setting.

An exception to this statement is provided by 
\cite{hoffman:2010:online, hoffman:2013:stochastic, wang:2011:online},
who have shown that a class of approximation 
methods known as \emph{variational Bayes} (VB)~\cite{wainwright:2008:graphical}
can be usefully deployed for large-scale data sets.  They have applied 
their approach,
referred to as \emph{stochastic variational inference} (SVI),
to the domain of topic modeling of document collections, 
an area with a major need for scalable inference algorithms.
VB traditionally uses the variational lower bound on the marginal
likelihood as an objective function, and the
idea of SVI is to apply a variant of stochastic gradient descent to 
this objective.  Notably, this objective is based 
on the conceptual existence of a full data set involving $D$ data points 
(i.e., documents in the topic model setting), for a fixed value of $D$.  
Although the stochastic gradient is computed for a single, small subset of data points
(documents) at a time, the posterior being targeted is a posterior for 
$D$ data points.  This value of $D$ must be specified in advance and is 
used by the algorithm at each step.  Posteriors for $D'$ data points, for 
$D' \ne D$, are not obtained as part of the analysis.  

We view this lack of a link between the number of documents that have
been processed thus far and the posterior that is being targeted as undesirable 
in many settings involving streaming data. In this paper we aim at 
an approximate Bayesian inference algorithm that is scalable like SVI
but is also truly a streaming procedure, in that it yields an approximate 
posterior for each processed collection of $D'$ data points---and not 
just a pre-specified ``final'' number of data points $D$.  
To that end, we return to the classical perspective of Bayesian 
updating, where the recursive application of Bayes theorem provides a 
sequence of posteriors, not a sequence of approximations to a fixed 
posterior.  To this classical recursive perspective we bring the 
VB framework; our updates need not be exact Bayesian updates but rather 
may be approximations such as VB.  This approach is similar in spirit to assumed 
density filtering or expectation propagation~\cite{minka:2001:expectation, 
minka:2001:family, opper:1998:bayesian}, but each step of those methods
involves a moment-matching step that can be computationally costly for models
such as topic models.  We are able to avoid the moment-matching step via 
the use of VB.  We also note other related work in this general vein: 
MCMC approximations have been explored by \cite{canini:2009:online},
and VB or VB-like approximations have also been explored by~\cite{honkela:2003:on-line, 
luts:2012:real}.

Although the empirical success of SVI is the main motivation for our
work, we are also motivated by recent developments in computer architectures,
which permit distributed and asynchronous computations in addition to streaming
computations.  As we will show, a streaming VB algorithm naturally lends
itself to distributed and asynchronous implementations.

\section{Streaming, distributed, asynchronous Bayesian updating} \label{sec:bayes_update}

\textbf{Streaming Bayesian updating.}
Consider data $x_{1}, x_{2}, \ldots$ generated iid according to a 
distribution $p(x \:|\: \Theta)$ given parameter(s) $\Theta$. Assume that a prior
$p(\Theta)$ has also been specified. Then Bayes theorem gives us 
the \emph{posterior distribution} of $\Theta$ given a collection 
of $S$ data points, $C_{1} := (x_{1}, \ldots, x_{S})$:
$$
	p(\Theta \mid C_{1}) = p(C_{1})^{-1} \; p(C_{1} \mid \Theta) \; p(\Theta),
$$
where
$
	p(C_{1} \:|\: \Theta)
		= p(x_{1}, \ldots, x_{S} \:|\: \Theta)
		= \prod_{s=1}^{S} p(x_{s} \:|\: \Theta).
$

Suppose we have seen and processed $b-1$ collections, sometimes called
\emph{minibatches}, of data.  Given the posterior $p(\Theta \:|\: C_{1}, 
\ldots, C_{b-1})$, we can calculate the posterior after the $b$th minibatch:
\begin{equation}
	\label{eq:old_post_as_new_prior}
	p(\Theta \mid C_{1}, \ldots, C_{b})
		\propto p(C_{b} \mid \Theta) \; p(\Theta \mid C_{1}, \ldots, C_{b-1}).
\end{equation}
That is, we treat the posterior after $b-1$ minibatches as the new prior for
the incoming data points. If we can save the posterior from $b-1$ minibatches
and calculate the normalizing constant for the $b$th posterior,
repeated application of \eq{old_post_as_new_prior} is streaming;
it automatically gives us the new posterior without
needing to revisit old data points.

In complex models, it is often infeasible to calculate the posterior
exactly, and an approximation must be used.
Suppose that, given a prior $p(\Theta)$ and data minibatch $C$,
we have an approximation algorithm
$\mathcal{A}$ that calculates an approximate posterior $q$:
$q(\Theta) = \mathcal{A}(C, p(\Theta))$. Then, setting
$q_{0}(\Theta) = p(\Theta)$, one way to recursively calculate an approximation to the posterior
is
\begin{equation}
	\label{eq:seq_bayes_update}
	p(\Theta \mid C_{1}, \ldots, C_{b})
		\approx q_{b}(\Theta)
		= \mathcal{A}\left( C_{b}, q_{b-1}(\Theta) \right).
\end{equation}
When $\mathcal{A}$ yields the posterior from Bayes theorem, this calculation is exact.
This approach already differs from that of
\cite{hoffman:2010:online, wang:2011:online, hoffman:2013:stochastic},
which we will see (\mysec{one_pass_lda})
directly approximates $p(\Theta \mid C_{1}, \ldots, C_{B})$
for fixed $B$
without making intermediate approximations for
$b$ strictly between $1$ and $B$.

\textbf{Distributed Bayesian updating.}
The sequential updates in \eq{seq_bayes_update}
handle streaming data in theory, but in practice, the $\mathcal{A}$
calculation might take longer than the time interval between
minibatch arrivals or simply take longer than desired.
Parallelizing computations
increases algorithm throughput.
And posterior calculations
need not be sequential. Indeed, 
Bayes theorem yields
\begin{align}
	\label{eq:parallel_theory}
	p(\Theta \mid C_{1}, \ldots, C_{B})
		&\propto \left[ \prod_{b=1}^{B} p(C_{b} \mid \Theta) \right] \; p(\Theta)
		\propto \left[ \prod_{b=1}^{B} p(\Theta \mid C_{b}) \; p(\Theta)^{-1} \right] p(\Theta).
\end{align}
That is, we can calculate the individual minibatch posteriors $p(\Theta \mid C_{b})$,
perhaps in parallel,
and then combine them to find the full posterior $p(\Theta \mid C_{1}, \ldots, C_{B})$.

Given an approximating algorithm $\mathcal{A}$ as above,
the corresponding approximate update would be
\begin{align}
	\label{eq:parallel_bayes_update}
	p(\Theta \mid C_{1}, \ldots, C_{B})
		&\approx q(\Theta)
		\propto \left[ \prod_{b=1}^{B} \mathcal{A}(C_{b}, p(\Theta)) \; p(\Theta)^{-1} \right]
			\; p(\Theta),
\end{align}
for some approximating distribution $q$, provided the normalizing constant 
for the right-hand side of \eq{parallel_bayes_update} can be computed.

Variational inference methods are generally based on exponential
family representations~\cite{wainwright:2008:graphical}, and we will
make that assumption here.  In particular, we suppose
$
	p(\Theta) \propto \exp\{ \xi_{0} \cdot T(\Theta) \};
$
that is, $p(\Theta)$ is an exponential family distribution for
$\Theta$ with sufficient statistic $T(\Theta)$ and natural parameter $\xi_{0}$.
We suppose further that $\mathcal{A}$ always returns a distribution in the
same exponential family; in particular, we suppose that there exists some
parameter $\xi_{b}$ such that
\begin{equation}
	\label{eq:A_exp_fam}
	q_{b}(\Theta) \propto \exp\{ \xi_{b} \cdot T(\Theta) \} \quad \textrm{for} \quad q_{b}(\Theta) = \mathcal{A}(C_{b}, p(\Theta)).
\end{equation}
When we make these two assumptions, the update in \eq{parallel_bayes_update}
becomes
\begin{equation}
	\label{eq:exp_fam_parallel_update}
	p(\Theta \mid C_{1}, \ldots, C_{B})
		\approx q(\Theta)
		\propto \exp\left\{ \left[ \xi_{0} + \sum_{b=1}^{B} (\xi_{b} - \xi_{0}) \right] \cdot T(\Theta) \right\},
\end{equation}
where the normalizing constant is readily obtained from the exponential
family form.  In what follows we 
use the shorthand $\xi \leftarrow \mathcal{A}(C, \xi_{0})$ to denote that 
$\mathcal{A}$ takes as input a minibatch $C$ and a prior with exponential family
parameter $\xi_{0}$ and that it returns a distribution in the same exponential family
with parameter $\xi$.

So, to approximate $p(\Theta \:|\: C_{1}, \ldots, C_{B})$, we first
calculate $\xi_{b}$ via the approximation primitive $\mathcal{A}$ for
each minibatch $C_{b}$; note that these calculations may be performed
in parallel. Then we sum together the quantities $\xi_{b} - \xi_{0}$
across $b$, along with the initial $\xi_{0}$ from the prior, to find
the final exponential family parameter to the full posterior approximation $q$.
We previously saw that the general Bayes sequential update can be made streaming
by iterating with the old posterior as the new prior (\eq{seq_bayes_update}).
Similarly, here we see that
the full posterior approximation $q$ is in the same exponential family as the prior, so 
one may iterate these parallel computations to arrive at a parallelized 
algorithm for streaming posterior computation.

We emphasize that while 
these updates are reminiscent of prior-posterior conjugacy, it is actually the 
approximate posteriors and single, original prior that we assume belong to the same exponential
family. It is not necessary to assume any conjugacy in the generative model
itself nor that any true intermediate or final posterior take any particular limited form.

\textbf{Asynchronous Bayesian updating.}
Performing $B$ computations in parallel can in theory
speed up algorithm running time by a factor of $B$, but in practice
it is often the case that a single computation thread takes longer than the rest.
Waiting for this thread to finish diminishes potential gains from
distributing the computations. This problem can be ameliorated
by making computations \emph{asynchronous}. In this case,
processors known as \emph{workers} each solve a subproblem.
When a worker finishes, it reports its solution to a single
\emph{master} processor. If the master gives the worker
a new subproblem without waiting for the other workers to finish,
it can decrease downtime in the system.

Our asynchronous algorithm is in the spirit of Hogwild!~\cite{niu:2011:hogwild}.
To present the algorithm we first describe an asynchronous computation 
that we will not use in practice, but which will serve as a conceptual
stepping stone.  Note in particular that the following scheme makes
the computations in \eq{exp_fam_parallel_update}
asynchronous. Have each worker continuously iterate between
three steps:
(1) collect a new minibatch $C$, (2) compute
the local approximate posterior $\xi \leftarrow \mathcal{A}(C, \xi_0)$, and (3) return
$\Delta \xi := \xi - \xi_{0}$ to the master.
The master, in turn, starts by assigning the posterior to equal the prior:
$\xi^{(\post)} \leftarrow \xi_0$.
Each time the master receives a quantity $\Delta \xi$
from any worker, it updates the posterior synchronously:
$\xi^{(\post)} \leftarrow \xi^{(\post)} + \Delta \xi$.
If $\mathcal{A}$ returns the exponential family parameter
of the true posterior (rather than an approximation), then the posterior at the master is exact
by \eq{parallel_bayes_update}.

A preferred asynchronous computation works as follows.
The master initializes its posterior estimate to the prior:
$\xi^{(\post)} \leftarrow \xi_0$.
Each worker continuously iterates between four steps:
(1) collect a new minibatch $C$,
(2) copy the master posterior value locally $\xi^{(\local)} \leftarrow \xi^{(\post)}$,
(3) compute the local approximate posterior $\xi \leftarrow \mathcal{A}(C, \xi^{(\local)})$,
and (4) return $\Delta \xi := \xi - \xi^{(\local)}$ to the master.
Each time the master receives a quantity $\Delta \xi$
from any worker, it updates the posterior synchronously:
$\xi^{(\post)} \leftarrow \xi^{(\post)} + \Delta \xi$.

The key difference between the first and second
frameworks proposed above
is that, in the second, the latest posterior
is used as a prior. This latter framework is more in line with the streaming
update of \eq{seq_bayes_update} but introduces a new layer
of approximation. Since $\xi^{(\post)}$ might change at the master
while the worker is computing $\Delta \xi$, it is no longer the case
that the posterior at the master is exact
when $\mathcal{A}$ returns the exponential family parameter
of the true posterior.
Nonetheless we find that the latter framework performs
better in practice, so we focus on it exclusively in what follows.

We refer to our overall framework as \emph{SDA-Bayes}, which stands
for (S)treaming, (D)istributed, (A)synchronous Bayes.  The framework
is intended to be general enough to allow a variety of local approximations
$\mathcal{A}$. Indeed, SDA-Bayes works
out of the box once an implementation of $\mathcal{A}$---and a prior
on the global parameter(s) $\Theta$---is provided.
In the current paper our preferred local approximation will
be VB.

\section{Case study: latent Dirichlet allocation} \label{sec:lda}

In what follows, we consider examples of
the choices for the $\Theta$ prior and primitive $\mathcal{A}$
in the context of \emph{latent Dirichlet allocation}
(LDA) \cite{blei:2003:latent}. 
LDA models the content of $D$ documents in a corpus. Themes potentially 
shared by multiple documents are described by \emph{topics}.
The unsupervised learning problem is to learn the topics as well 
as discover which topics occur in which documents. 

More formally, each topic (of $K$ total topics) is a distribution over the $V$ words
in the vocabulary: $\beta_{k} = (\beta_{kv})_{v=1}^{V}$. Each 
document is an admixture of topics. The words in document $d$
are assumed to be exchangeable. Each word $w_{dn}$
belongs to a latent topic $z_{dn}$ chosen according to a document-specific
distribution of topics $\theta_{d} = (\theta_{dk})_{k=1}^{K}$. The
full generative model, with Dirichlet priors for $\beta_{k}$ and $\theta_{d}$
conditioned on respective parameters $\eta_{k}$ and $\alpha$,
appears in \cite{blei:2003:latent}.

To see that this model fits our specification in \mysec{bayes_update},
consider the set of global parameters $\Theta = \beta$. 
Each document $w_{d} = (w_{dn})_{n=1}^{N_{d}}$ is distributed
iid conditioned on the global topics. The full collection of 
data is a corpus $C = w = (w_{d})_{d=1}^{D}$ of documents.
The posterior for LDA, $p(\beta, \theta, z \mid C, \eta, \alpha)$, is
equal to the following expression up to proportionality:
\begin{equation}
	\label{eq:lda_posterior}
		\propto \left[ \prod_{k=1}^{K} \dir(\beta_{k} \mid \eta_{k}) \right]
			\cdot \left[ \prod_{d=1}^{D} \dir(\theta_{d} \mid \alpha) \right] \\
			\cdot \left[ \prod_{d=1}^{D} \prod_{n=1}^{N_{d}} \theta_{dz_{dn}} \beta_{z_{dn}, w_{dn}} \right].
\end{equation}
The posterior for just the global parameters $p(\beta \:|\: C, \eta, \alpha)$ can be obtained
from $p(\beta, \theta, z \:|\: C, \eta, \alpha)$ by integrating out the local, document-specific 
parameters $\theta, z$.
As is common in complex models, the normalizing constant for \eq{lda_posterior}
is intractable to compute, so the posterior must be approximated.

\subsection{Posterior-approximation algorithms} \label{sec:approx_lda}

To apply SDA-Bayes to LDA, we use the prior specified
by the generative model.
It remains to choose a posterior-approximation algorithm $\mathcal{A}$. 
We consider two possibilities here:
variational Bayes (VB) and expectation propagation (EP). 
Both primitives take Dirichlet distributions as priors for $\beta$
and both return
Dirichlet distributions for the approximate posterior of the topic parameters
$\beta$; thus the prior and approximate posterior are in the same exponential family.
Hence both VB and EP can be utilized as a choice for $\mathcal{A}$ in the SDA-Bayes
framework. 

\newcommand{\BatchVB}{
	\begin{algorithm}[H]
		\SetKwFunction{locvars}{LocalVB}
		\KwIn{Data $(n_{d})_{d=1}^{D}$; hyperparameters $\eta, \alpha$}
		\KwOut{$\lambda$}
		Initialize $\lambda$ \\
		\While{$(\lambda, \gamma, \phi)$ not converged}{
			\For{$d = 1,\dots,D$}{
				$(\gamma_d, \phi_d) \leftarrow$ \locvars{$d, \lambda$}
			}
			$\forall (k,v)$, $\lambda_{kv} \leftarrow \eta_{kv} + \sum_{d=1}^D \phi_{dvk} n_{dv}$
		}
	\caption{\label{alg:batchvb_lda} VB for LDA} 
	\end{algorithm}
}

\newcommand{\LocUpdate}{
	\RestyleAlgo{tworuled}
	\SetAlgorithmName{Subroutine}{subroutine}{List of Subroutines}
	\begin{algorithm}[H]
		\SetKwFunction{locvars}{LocalVB}
		\SetKwProg{subr}{Subroutine}{}{}
		\subr{\locvars{$d, \lambda$}}{	
			\KwOut{$(\gamma_{d}, \phi_{d})$}
			Initialize $\gamma_{d}$ \\
			\While{$(\gamma_d, \phi_d)$ not converged}{
				$\forall (k,v)$, set
					$\phi_{dvk} \propto \exp \left(
						\mbe_{q}[\log \theta_{dk}] + \mbe_{q}[\log \beta_{kv}]
					\right)$ (normalized across $k$) \\
				$\forall k$, $\gamma_{dk} \leftarrow \alpha_{k} + \sum_{v=1}^{V} \phi_{dvk} n_{dv}$
			}
		}
	\end{algorithm}
}

\newcommand{\SSAlg}{
	\begin{algorithm}[H]
	\SetKwFunction{locvars}{LocalVB}
	\KwIn{Hyperparameters $\eta, \alpha$}
	\KwOut{A sequence $\lambda^{(1)}, \lambda^{(2)}, \ldots$}
	Initialize $\forall (k,v)$, $\lambda^{(0)}_{kv} \leftarrow \eta_{kv}$ \\
	\For{$b = 1, 2, \ldots$}{
		Collect new data minibatch $C$ \\
		\ForEach{ document indexed $d$ in $C$ }{
			$(\gamma_{d}, \phi_{d}) \leftarrow$ \locvars{$d, \lambda$}
		}
		$\forall (k,v)$, $\lambda^{(b)}_{kv} \leftarrow \lambda^{(b-1)}_{kv} + \sum_{d \textrm{ in } C} \phi_{dvk} n_{dv}$
	}
	\caption{\label{alg:ss_lda} SSU for LDA}
	\end{algorithm}
}

\newcommand{\SVI}{
	\begin{algorithm}[H]
		\SetKwFunction{locvars}{LocalVB}
		\KwIn{Hyperparameters $\eta, \alpha, D, (\rho_{t})_{t=1}^{T}$}
		\KwOut{$\lambda$}
		Initialize $\lambda$ \\
		\For{$t=1,\ldots,T$}{
			Collect new data minibatch $C$ \\
			\ForEach{ document indexed $d$ in $C$ }{
				$(\gamma_{d}, \phi_{d}) \leftarrow$ \locvars{$d, \lambda$}
			}
			$\forall (k,v)$, $\tilde{\lambda}_{kv} \leftarrow \eta_{kv} + \frac{D}{|C|} \sum_{d \textrm{ in } C} \phi_{dvk} n_{dv}$ \\
			$\forall (k,v)$, $\lambda_{kv} \leftarrow (1-\rho_{t}) \lambda_{kv} + \rho_{t} \tilde{\lambda}_{kv}$
		}
	\caption{\label{alg:svi} SVI for LDA}
	\end{algorithm}
}

\begin{figure}
	\centering
		%
		\BatchVB 
		
		\LocUpdate 
		
		%
		%
		\SVI 
		
		\SSAlg
		%
	\caption{\label{fig:algs_vb} Algorithms for calculating $\lambda$, the 
	parameters for the topic posteriors in LDA. VB iterates multiple times
	through the data, SVI makes a single pass, and SSU is streaming. Here,
	$n_{dv}$ represents the number of words $v$ in document $d$.
	}
	\vspace{-0.5cm}
\end{figure}

\textbf{Mean-field variational Bayes.}
We use the shorthand $p_{D}$ for \eq{lda_posterior}, the posterior
given $D$ documents.
We assume the approximating distribution, written $q_{D}$ for shorthand, takes the form
\begin{align}
	\nonumber
	\lefteqn{ q_{D}(\beta, \theta, z \mid \lambda, \gamma, \phi) } \\
		\label{eq:lda_vb_approx_q}
		&= \left[ \prod_{k=1}^{K} q_{D}(\beta_{k} \mid \lambda_{k})
			\right]
			\cdot \left[ \prod_{d=1}^{D} q_{D}(\theta_{d} \mid \gamma_{d}) \right]
			\cdot \left[ \prod_{d=1}^{D} \prod_{n=1}^{N_{d}} q_{D}(z_{dn} \mid \phi_{dw_{dn}}) \right]
\end{align}
for parameters $(\lambda_{kv}), (\gamma_{dk}), (\phi_{dvk})$
with $k \in \{1,\ldots,K\}, v \in \{1,\ldots,V\}, d\in \{1,\ldots,D\}$.
Moreover, we set $q_{D}(\beta_{k} \:|\: \lambda_{k}) = \dir_{V}(\beta_{k} \:|\: \lambda_{k})$,
$q_{D}(\theta_{d} \:|\: \gamma_{d}) = \dir_{K}(\theta_{d} \:|\: \gamma_{d})$, and
$q_{D}(z_{dn} \:|\: \phi_{dw_{dn}}) = \cat_{K}(z_{dn} \:|\: \phi_{dw_{dn}})$. The subscripts
on $\dir$ and $\cat$ indicate the dimensions
of the distributions (and of the parameters).

The problem of VB is to find the best approximating $q_{D}$, defined as the 
collection of variational parameters $\lambda, \gamma, \phi$
that minimize the KL divergence from the true posterior: $\dkl{q_{D}}{p_{D}}$.
Even finding the minimizing parameters is a difficult optimization problem.
Typically the solution
is approximated by coordinate descent in each parameter
\citep{blei:2003:latent,wainwright:2008:graphical} as in \alg{batchvb_lda}.
The derivation
of VB for LDA can be found in \citep{blei:2003:latent,hoffman:2013:stochastic}
and \app{batch_vb}.

\textbf{Expectation propagation.}
An EP \cite{minka:2001:expectation}
algorithm for approximating
the LDA posterior appears in \alg{ep_lda} of \app{ep}.
\alg{ep_lda} differs from \cite{minka:2002:expectation},
which does not provide an approximate posterior for the topic
parameters, and is instead our own derivation.
Our version of EP, like VB, learns factorized Dirichlet 
distributions over topics.

\subsection{Other single-pass algorithms for approximate LDA posteriors} \label{sec:one_pass_lda}

The algorithms in \mysec{approx_lda} pass through the
data multiple times and require storing the data set
in memory---but are useful as primitives for SDA-Bayes
in the context of the processing of minibatches of data.
Next, we consider two algorithms that can pass through
a data set just one time (\emph{single pass}) and to 
which we compare in the evaluations (\mysec{evaluation}).

\textbf{Stochastic variational inference.}
VB uses coordinate descent to
find a value of $q_{D}$, \eq{lda_vb_approx_q}, that
locally minimizes the KL
divergence, $\dkl{q_{D}}{p_{D}}$.
\emph{Stochastic variational
inference} (SVI) \citep{hoffman:2010:online,hoffman:2013:stochastic}
is exactly the application of a particular version of stochastic gradient 
descent
to the same optimization problem. 
While stochastic gradient descent can often be viewed as a streaming algorithm,
the optimization problem itself here depends on $D$ via $p_{D}$, the posterior 
on $D$ data points. We see that, as a result, $D$ must be specified in advance,
appears in each step of SVI (see \alg{svi}), and is
independent of the number of data points actually processed by the
algorithm. Nonetheless, while one may choose to visit
$D' \ne D$ data points or 
revisit data points when using SVI to estimate $p_{D}$
\citep{hoffman:2010:online,hoffman:2013:stochastic},
SVI can be made single-pass by visiting each of
$D$ data points exactly once
and then
has constant memory requirements.
We also note that two new parameters,
$\tau_{0} > 0$ and $\kappa \in (0.5,1]$,
appear in SVI, beyond those in VB,
to determine a learning rate $\rho_{t}$
as a function of iteration $t$:
$\rho_{t} := (\tau_{0} + t)^{-\kappa}$.

\textbf{Sufficient statistics.}
On each round of VB (\alg{batchvb_lda}), we update the local
parameters for all documents and then compute
$\lambda_{kv} \leftarrow \eta_{kv} + \sum_{d=1}^D \phi_{dvk} n_{dv}$.
An alternative single-pass (and indeed streaming) option
would be to update the local parameters for
each minibatch of documents as they arrive and then
add the corresponding terms $\phi_{dvk} n_{dv}$ to the current
estimate of $\lambda$ for each document $d$ in the minibatch.
This essential idea has been proposed previously for models 
other than LDA by \cite{honkela:2003:on-line,luts:2012:real}
and forms the basis of what we call the
\emph{sufficient statistics update algorithm} (SSU): \alg{ss_lda}.
This algorithm is equivalent to SDA-Bayes with $\mathcal{A}$
chosen to be a single iteration over the global
variable $\lambda$ of VB (i.e., updating $\lambda$ exactly
once instead of iterating until convergence).

\section{Evaluation} \label{sec:evaluation}


We follow \cite{hoffman:2013:stochastic} (and further
\cite{teh:2006:collapsed, asuncion:2009:smoothing})
in evaluating our algorithms
by computing (approximate) predictive probability.
Under this metric, a higher score is better,
as a better model will assign a higher probability to the held-out words.

We calculate predictive probability
by first setting aside held-out testing documents $\ctest$
from the full corpus and then further setting aside a subset of
held-out testing words $\wdtest$ in each testing document $d$. 
The
remaining (training)
documents $\ctrain$ are used to estimate the global parameter posterior $q(\beta)$,
and the remaining (training) words $\wdtrain$ within the $d$th testing document are used
to estimate the document-specific parameter posterior $q(\theta_d)$.\footnote{
In all cases, we estimate $q(\theta_d)$ for evaluative purposes using VB
since direct EP estimation takes prohibitively long.}
To calculate predictive probability,
an approximation is necessary since we do not know the
predictive distribution---just as we seek to learn the posterior
distribution.
Specifically, we calculate the normalized predictive distribution
and report ``log predictive probability'' as
\begin{align*}
	\lefteqn{
	\frac{
		\sum_{d \in \ctest}
		\log p(\wdtest \mid \ctrain, \wdtrain)
	}{
		\sum_{d \in \ctest} \left| \wdtest \right|
	}
	} \\
	& = \frac{
		\sum_{d \in \ctest} \sum_{w_\test \in \wdtest}
		\log p(w_\test \mid \ctrain, \wdtrain)
	}{
		\sum_{d \in \ctest} \left| \wdtest \right|
	},
\end{align*}
where we use the approximation
\begin{align*}
	\lefteqn{ p(w_\test \mid \ctrain, \wdtrain) } \\
		&= \int_\beta \int_{\theta_{d}}
				\left(\sum_{k=1}^K \theta_{dk} \beta_{kw_\test}\right)
				p(\theta_{d} \mid W_{d, \train}, \beta) \: p(\beta \mid C^{(\train)})
			\: d\theta_{d} \: d\beta \\
		&\approx  \int_\beta \int_{\theta_{d}}
				\left(\sum_{k=1}^K \theta_{dk} \beta_{kw_\test} \right)  q(\theta_{d}) \: q(\beta)
			\: d\theta_{d} \: d\beta
		= \sum_{k=1}^K \mathbb{E}_q[\theta_{dk}] \: \mathbb{E}_q[\beta_{kw_\test}].
\end{align*}


To facilitate comparison with SVI, we use
the Wikipedia and Nature corpora of
\cite{hoffman:2010:online, wang:2011:online} in our experiments. 
These two corpora represent a range of sizes
(3{,}611{,}558 training documents for Wikipedia and 351{,}525 for Nature) as well as 
different types of topics. We expect words in Wikipedia to represent
an extremely broad range of topics
whereas we expect words in Nature to focus more
on the sciences.
We further use the vocabularies of \cite{hoffman:2010:online, wang:2011:online}
and SVI code available online at \cite{hoffman:2010:online:code}.
We hold out $10{,}000$ Wikipedia documents
and $1{,}024$ Nature documents (not
included in the counts above) for testing.
In the results presented in the main text, we follow
\cite{hoffman:2010:online,hoffman:2013:stochastic}
in
fitting an LDA model with $K=100$ topics
and hyperparameters chosen as:
$\forall k, \alpha_{k} = 1/K$, $\forall (k,v), \eta_{kv} = 0.01$.
For both Wikipedia and Nature, we set the parameters in SVI
according to the optimal values of the parameters described in Table 1 
of \cite{hoffman:2010:online} (number of documents $D$
correctly set in advance,
step size parameters $\kappa = 0.5$ and $\tau_0 = 64$).

\figs{wiki_minibatch} and \figss{nature_minibatch}
demonstrate that both SVI and SDA are sensitive to 
minibatch size when $\eta_{kv}=0.01$,
with generally superior performance 
at larger batch sizes. 
Interestingly, both SVI and SDA performance
improve and are steady across batch size
when $\eta_{kv}=1$ (\figs{wiki_minibatch} and \figss{nature_minibatch}).
Nonetheless, we use $\eta_{kv}=0.01$ in what follows
in the interest of consistency with
\cite{hoffman:2010:online,hoffman:2013:stochastic}.
Moreover, in the remaining experiments, 
we use a large minibatch size of $2^{15} = 32{,}768$.
This size is the largest before SVI performance
degrades in the Nature data set (\fig{nature_minibatch}).

\begin{table}
	\newcolumntype{.}{D{.}{.}{-1}} 
	\makebox[\textwidth][c]{
	\begin{tabular}{r........}
	\toprule
	& \multicolumn{4}{c}{Wikipedia}
	& \multicolumn{4}{c}{Nature} \\
		\cmidrule(r){2-5} \cmidrule(r){6-9}
		& \multicolumn{1}{c}{32-SDA} & \multicolumn{1}{c}{1-SDA} & \multicolumn{1}{c}{SVI} & \multicolumn{1}{c}{SSU}
		& \multicolumn{1}{c}{32-SDA} & \multicolumn{1}{c}{1-SDA} & \multicolumn{1}{c}{SVI} & \multicolumn{1}{c}{SSU} \\
		\cmidrule(r){2-5} \cmidrule(r){6-9}
	Log pred prob
		& -\textbf{7}.\textbf{31}
		& -7.43 & -7.32 & -7.91
		& -7.11
		& -7.19 & -\textbf{7}.\textbf{08} & -7.82
		\\
	Time (hours)
		& \textbf{2}.\textbf{09} 
		& 43.93 & 7.87 & 8.28
		& \textbf{0}.\textbf{55}
		& 10.02 & 1.22 & 1.27
		\\
	\bottomrule
	\end{tabular}
	}
	\caption{\label{tab:best_svi_perf} A comparison of (1) log predictive probability of held-out data
	and (2) running time of four algorithms: SDA-Bayes with 32 threads, SDA-Bayes with 1 thread,
	SVI, and SSU.}
\end{table}

Performance and
timing results are shown in \tab{best_svi_perf}.
One would expect that with additional streaming capabilities, SDA-Bayes
should show a performance loss relative to SVI. We see from \tab{best_svi_perf}
that such loss is small in the single-thread case,
while SSU performs much worse.  SVI is faster than single-thread SDA-Bayes
in this single-pass setting.

\textbf{Full SDA-Bayes improves run time with no performance cost.}
We handicap SDA-Bayes in the above comparisons
by utilizing just a single thread. In \tab{best_svi_perf}, we
also report performance of SDA-Bayes with 32 threads 
and the same minibatch size. In the synchronous case, we consider
minibatch size to equal the total number of data points
processed per round; therefore, the minibatch size
equals the number of data points sent to each
thread per round times the total number of threads.
In the asynchronous case, we analogously report minibatch 
size as this product.

\begin{figure}
	\begin{center}
	
	\subfigure[Wikipedia]{
		\includegraphics[width=.45\textwidth]{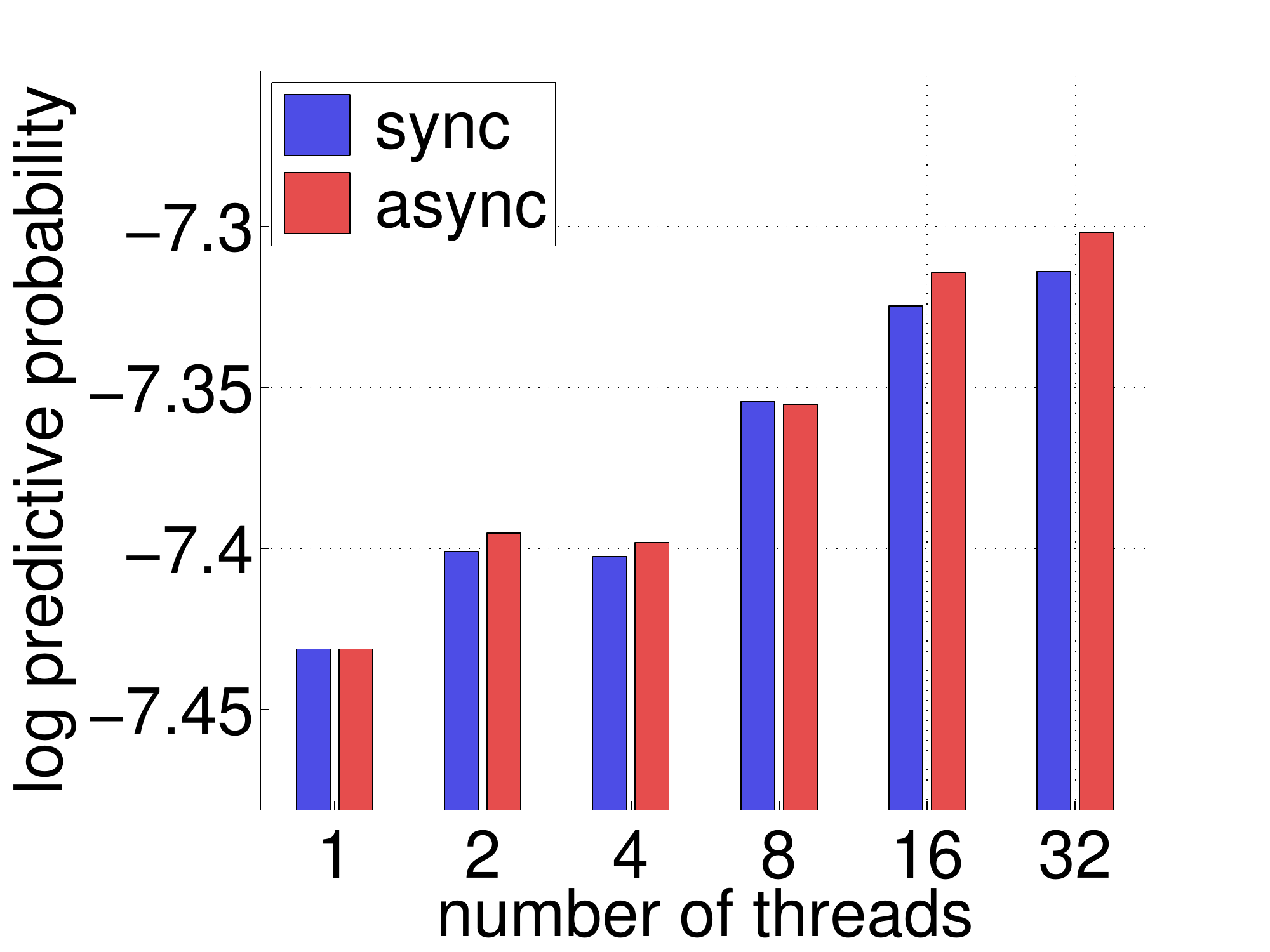}
		\label{fig:wiki_distr_perf}
	}
	\subfigure[Nature]{
		\includegraphics[width=.45\textwidth]{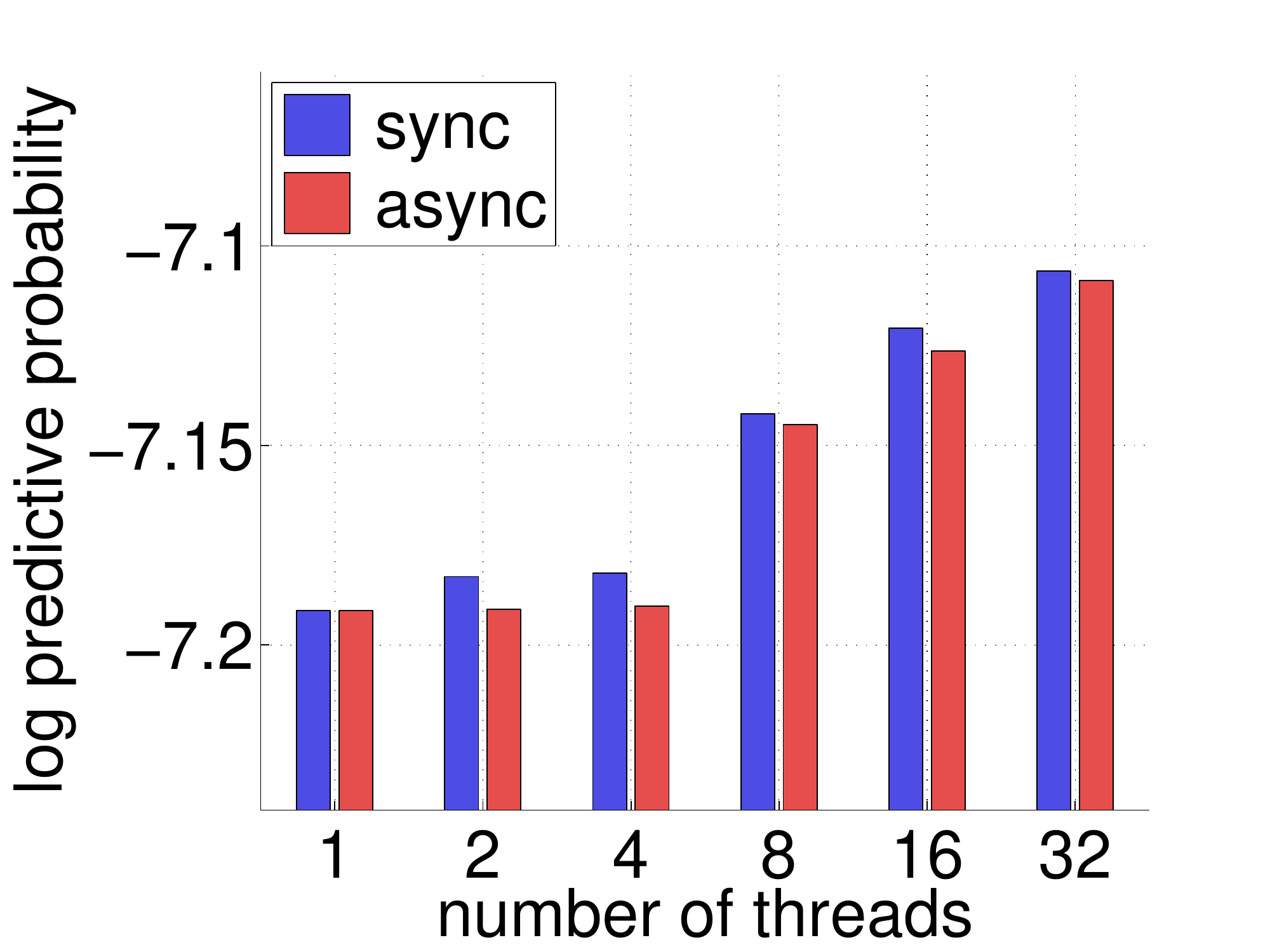}
		\label{fig:nature_distr_perf}
	}
	
	\subfigure[Wikipedia]{
		\includegraphics[width=.45\textwidth]{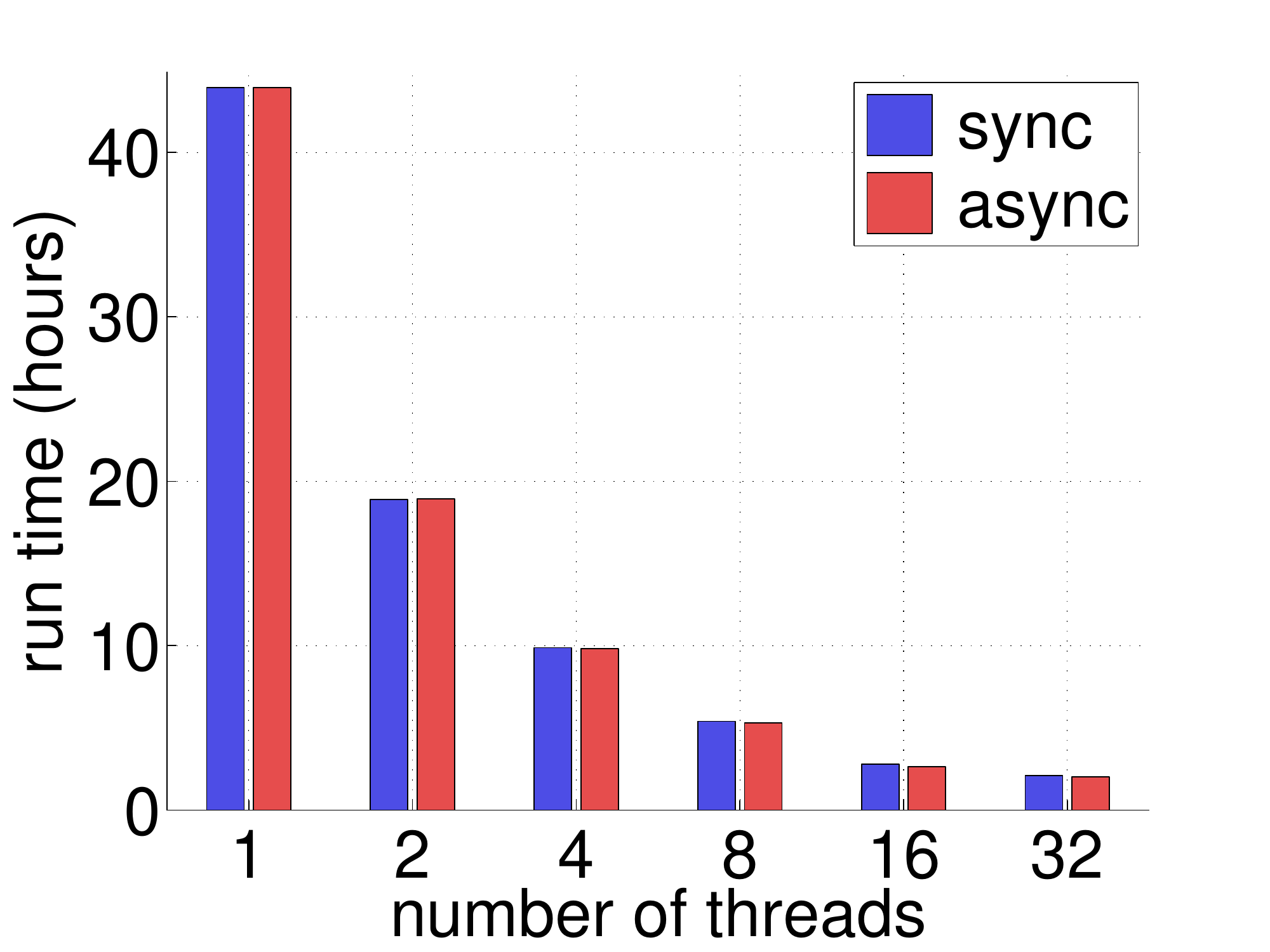}
		\label{fig:wiki_distr_time}
	}
	\subfigure[Nature]{
		\includegraphics[width=.45\textwidth]{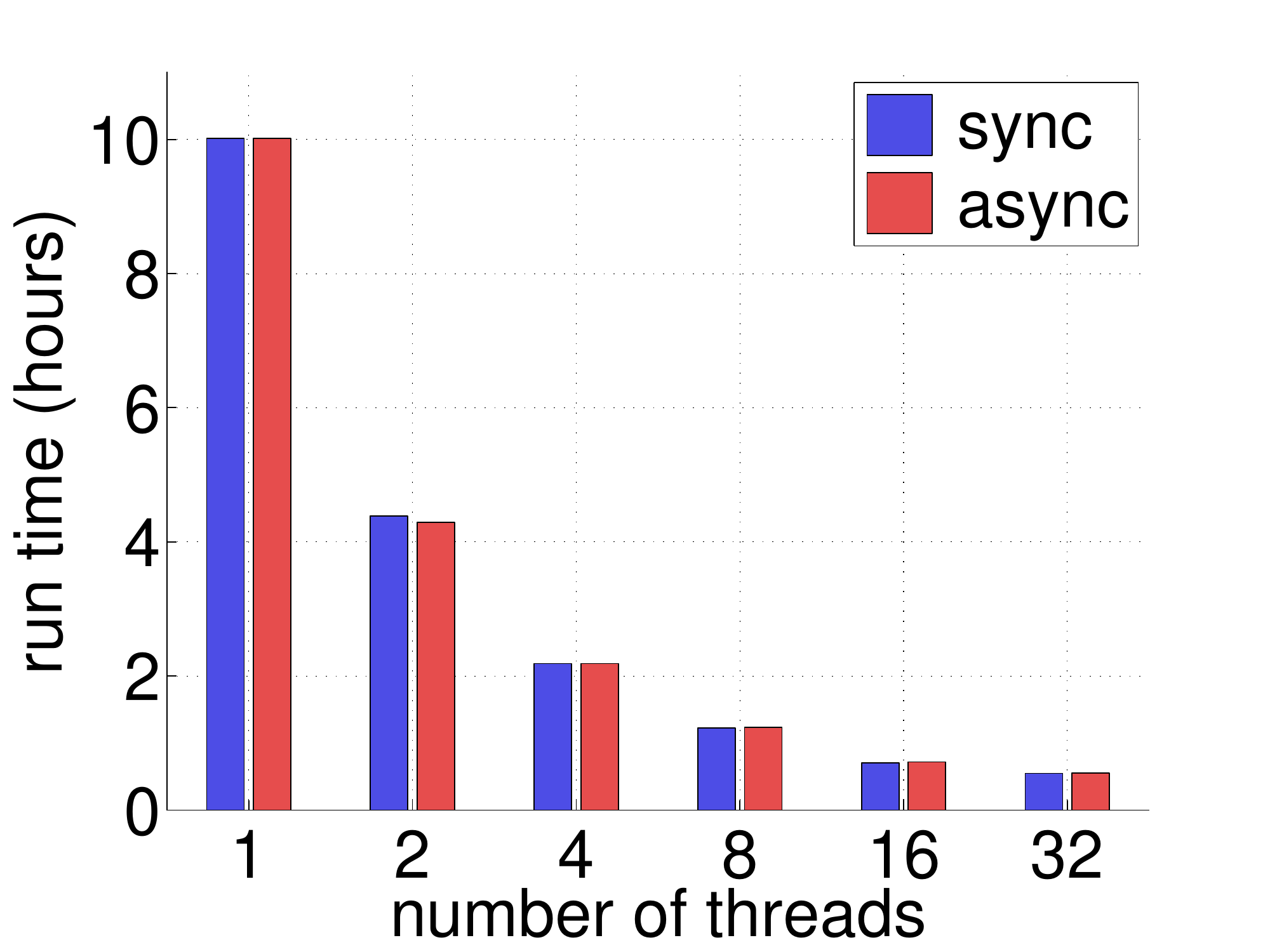}
		\label{fig:nature_distr_time}
	}
	
	\end{center}
	\caption{\label{fig:distr} SDA-Bayes log predictive probability (\emph{two upper plots}) and run time (\emph{two lower plots}) as a function of number of threads. 
	}
\end{figure}

\fig{distr} shows the performance of SDA-Bayes when we run with
$\{1,2,4,8,16,32\}$ threads while keeping the minibatch size constant.
The goal in such a distributed context is to improve run time while
not hurting performance. Indeed, we see dramatic run time improvement
as the number of threads grows
and in fact some slight performance improvement as well.
We tried both a parallel version and a full distributed, asynchronous version
of the algorithm; \fig{distr} indicates that the speedup and 
performance improvements we see here
come from parallelizing---which is theoretically justified
by \eq{parallel_theory} when $\mathcal{A}$
is Bayes rule. Our experiments indicate that our Hogwild!-style
asynchrony 
does not hurt performance. In our experiments,
the processing time at each thread seems to be
approximately equal across threads and dominate any
communication time at the master, so synchronous and
asynchronous performance and running time are essentially
identical.
In general, a practitioner might
prefer asynchrony since it is more robust to node failures.

\textbf{SVI is sensitive to the choice of total data size $D$.}
The evaluations above are for a single posterior over $D$ data points.
Of greater concern to us in this work is the evaluation of algorithms
in the streaming setting.
We have seen that SVI is designed to find the posterior for a
particular, pre-chosen number of data points $D$. 
In practice, when we 
run SVI on the full data set but change the input value of
$D$ in the algorithm, we can see
degradations in performance. In particular, we try values of 
$D$ equal to $\{0.01, 0.1, 1, 10, 100\}$ times the true $D$ in
\fig{wiki_svi_D} for the Wikipedia data set
and in
\fig{nature_svi_D} for the Nature data set.

A practitioner in the streaming setting will typically not know
$D$ in advance, or multiple values of $D$ may be of interest.
\figs{wiki_svi_D} and \figss{nature_svi_D} illustrate that an estimate
may not be sufficient. Even in the case where $D$ is known in
advance, it is reasonable to imagine a new influx of further data.
One might need to run SVI again
from the start (and, in so doing, revisit the first data set)
to obtain the desired performance.

\begin{figure}
	\hfill
	\subfigure[Sensitivity to minibatch size on Wikipedia]{
		\includegraphics[width=.45\textwidth]{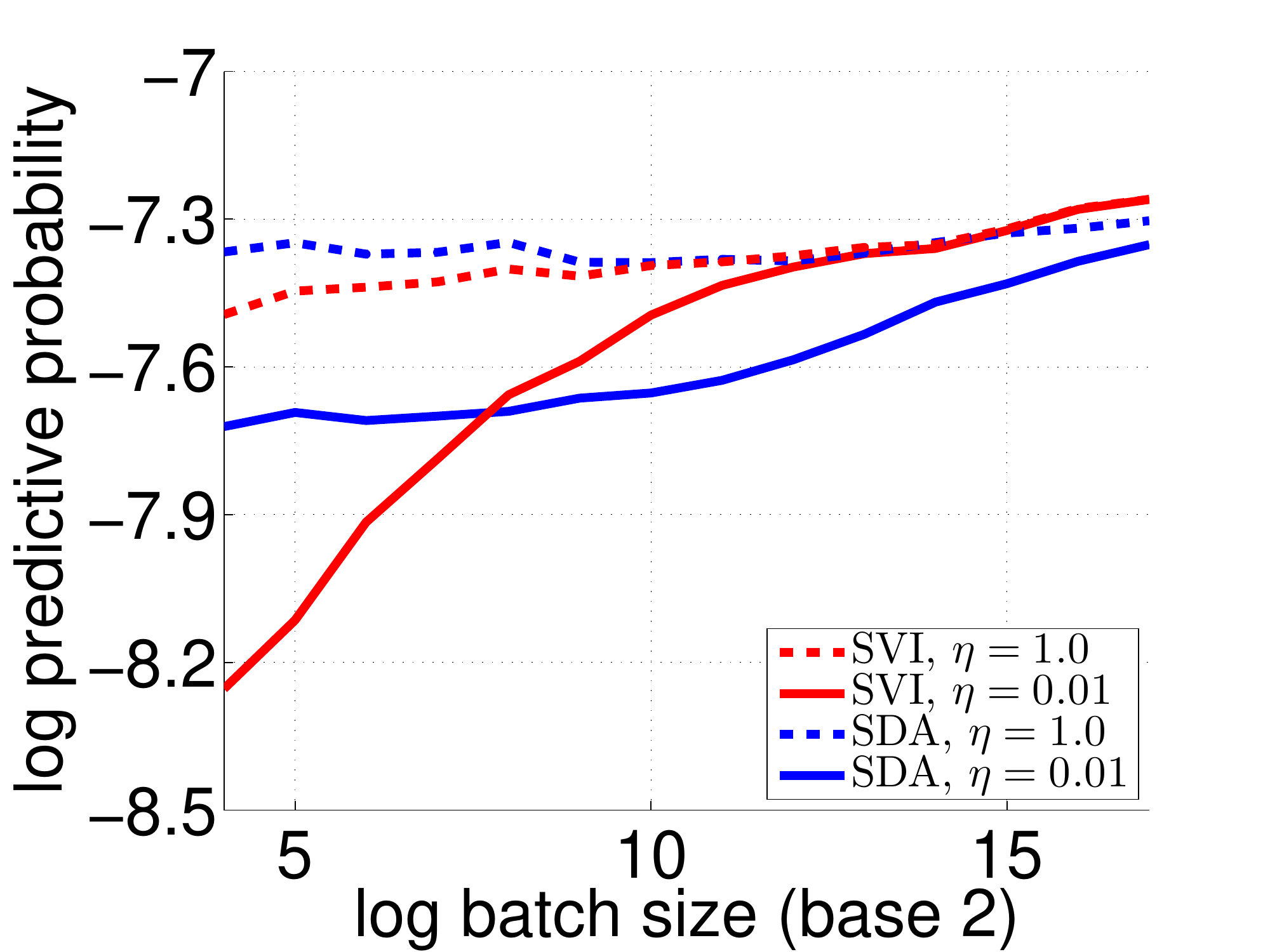}
		\label{fig:wiki_minibatch}
	}
	\hfill
	\subfigure[Sensitivity to minibatch size on Nature]{
		\includegraphics[width=.45\textwidth]{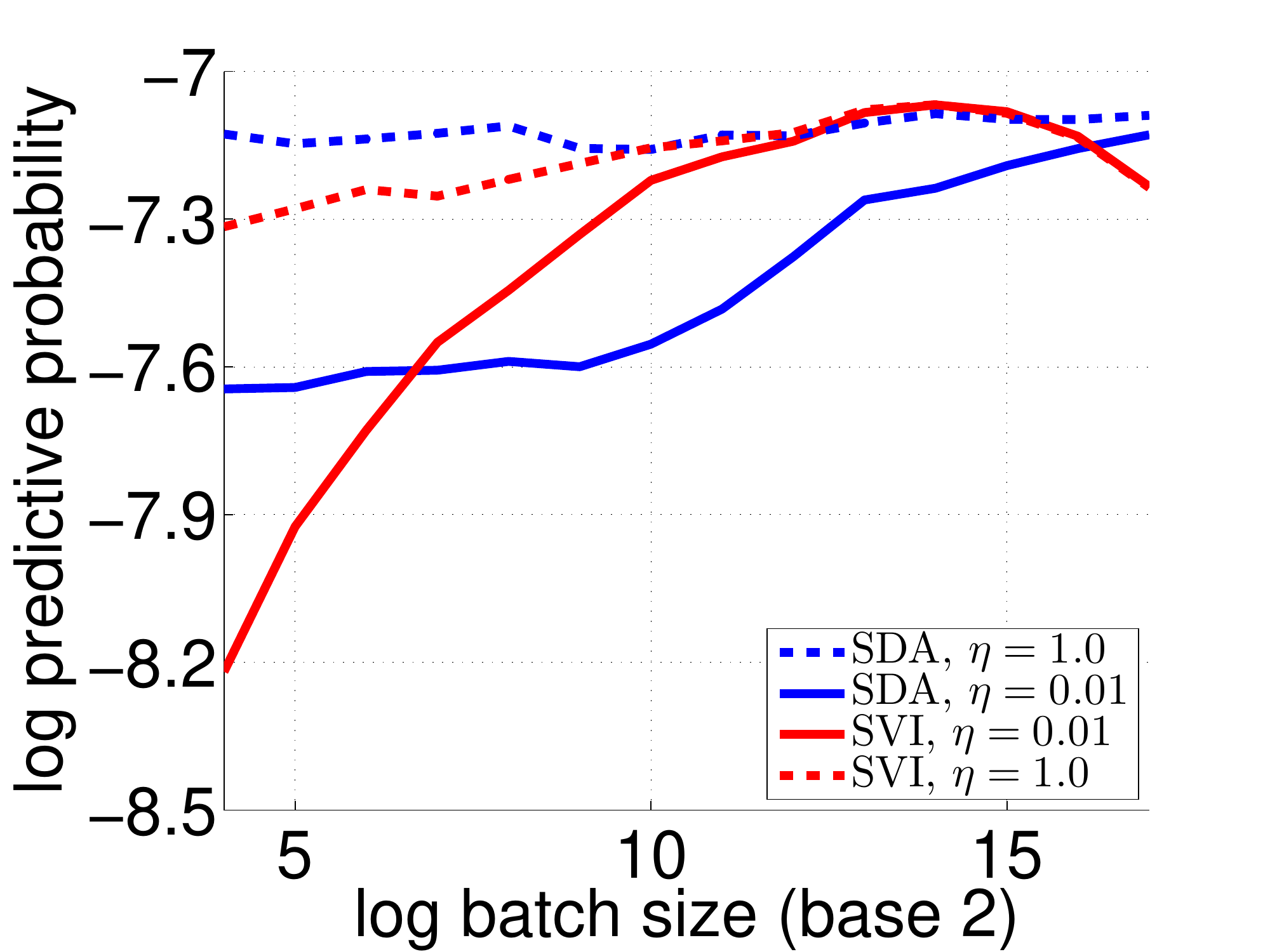}
		\label{fig:nature_minibatch}
	}
	\hfill
	\\
	\subfigure[SVI sensitivity to $D$ on Wikipedia]{
		\includegraphics[width=.45\textwidth]{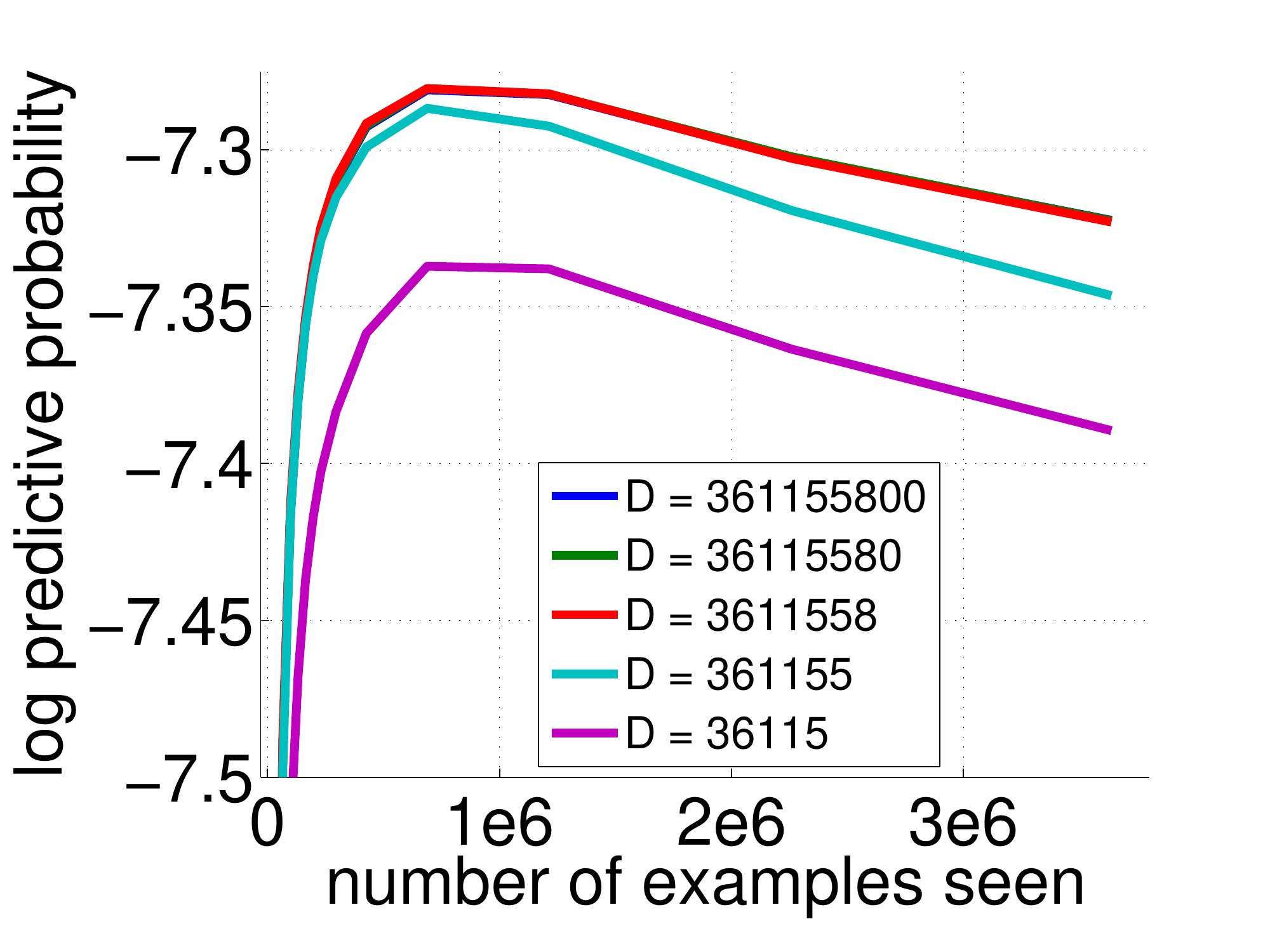}
		\label{fig:wiki_svi_D}
	}
	\hfill
	\subfigure[SVI sensitivity to $D$ on Nature]{
		\includegraphics[width=.45\textwidth]{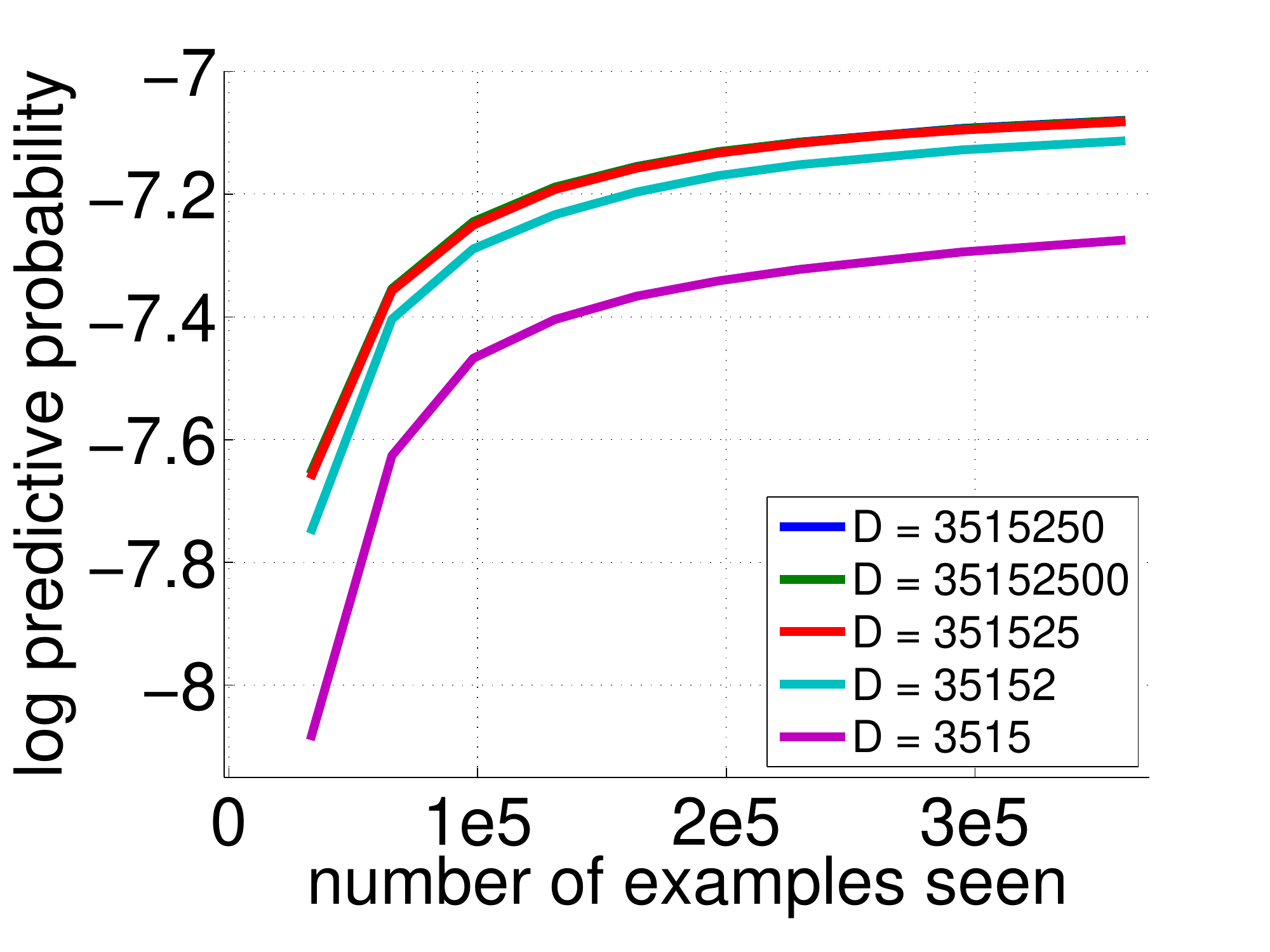}
		\label{fig:nature_svi_D}
	}
	\hfill
	\\
	\subfigure[SVI sensitivity to stepsize parameters on Wikipedia]{
		\includegraphics[width=.45\textwidth]{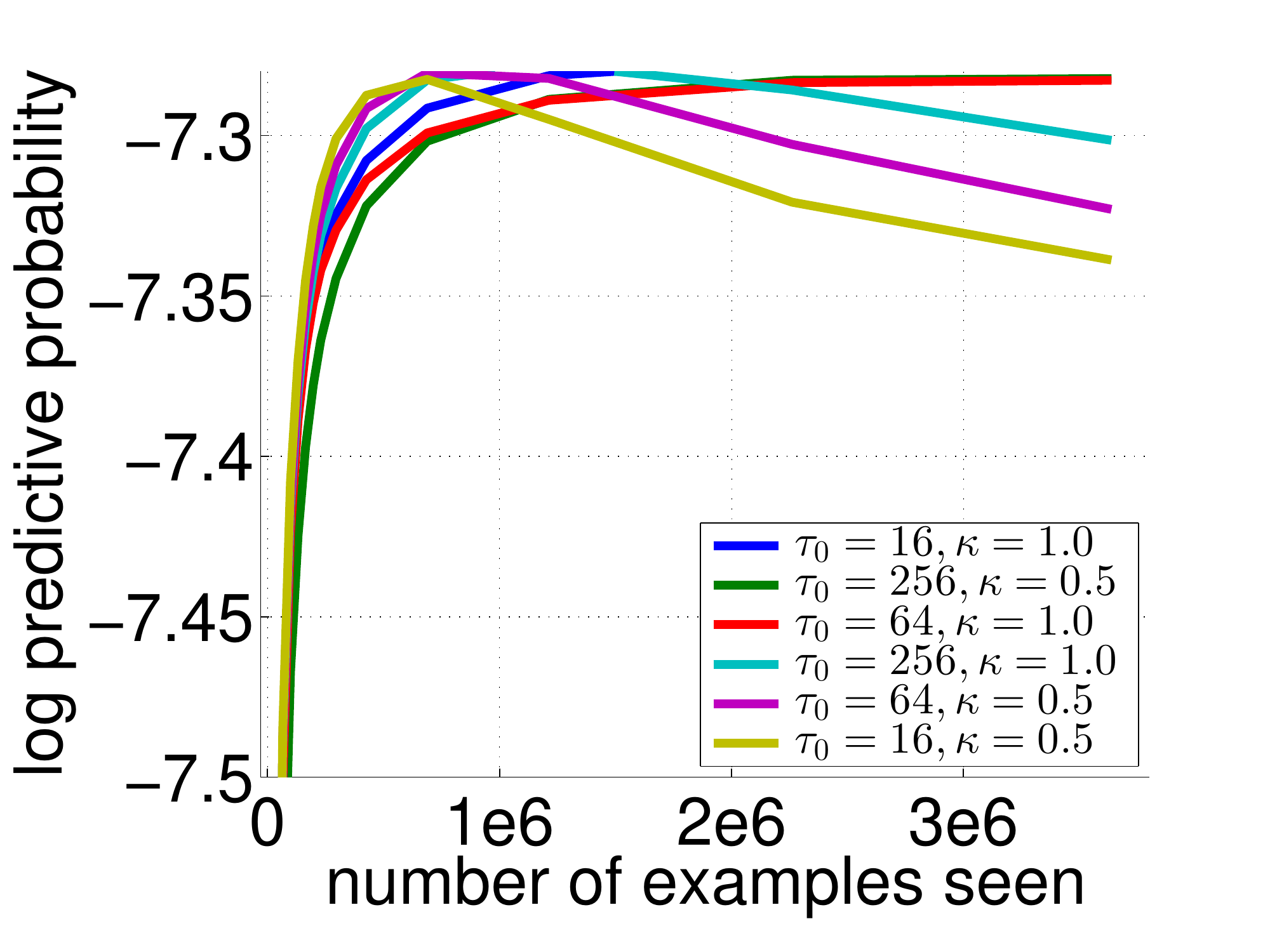}
		\label{fig:wiki_svi_step_size}
	}
	\hfill
	\subfigure[SVI sensitivity to stepsize parameters on Nature]{
		\includegraphics[width=.45\textwidth]{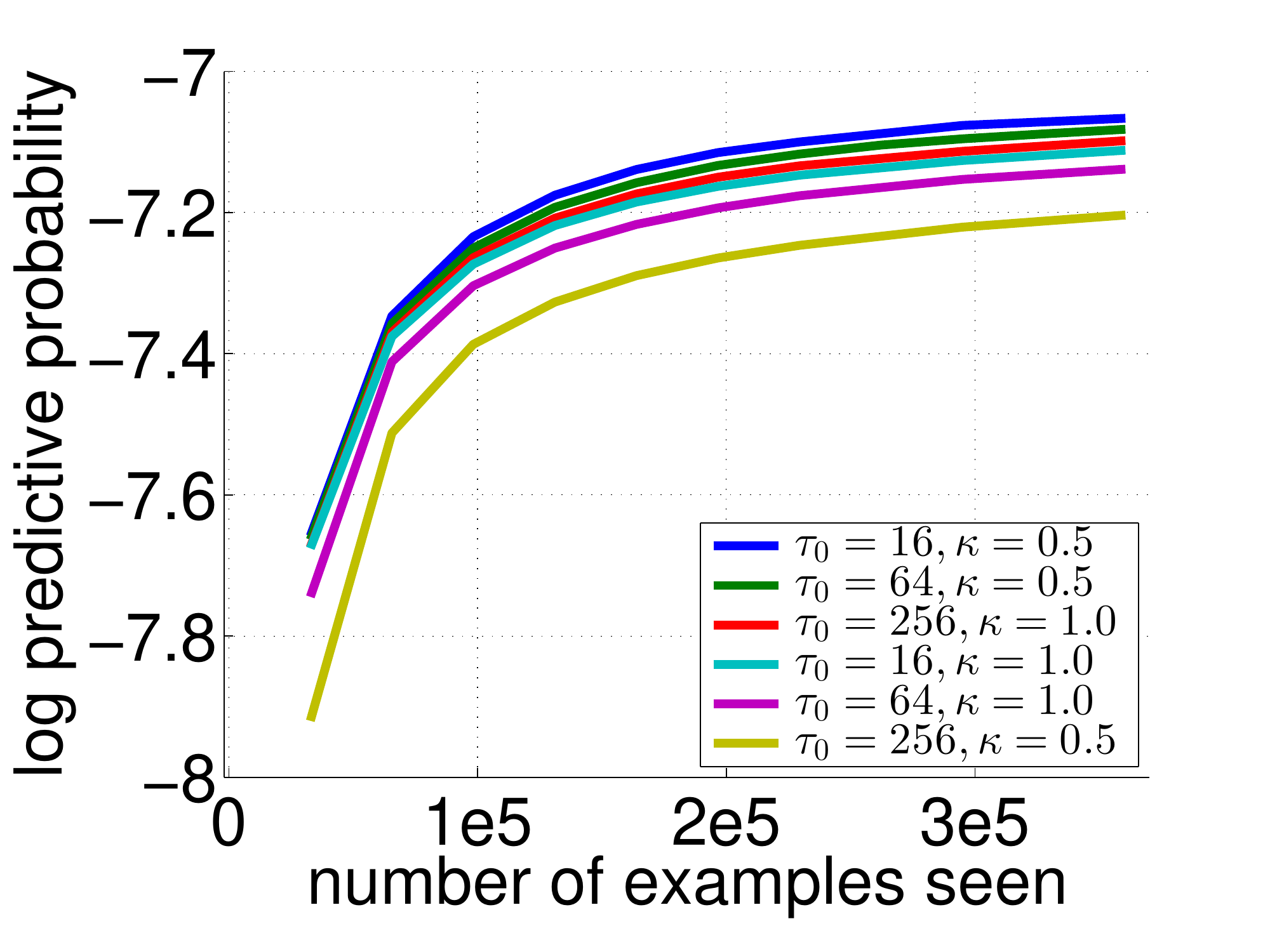}
		\label{fig:nature_svi_step_size}
	}
	\hfill
	\caption{\label{fig:eval_params} Sensitivity of SVI and SDA-Bayes to some respective parameters. Legends have the same top-to-bottom order as the rightmost curve points.}
\end{figure}

\textbf{SVI is sensitive to learning step size.}
\cite{hoffman:2010:online, wang:2011:online} use cross-validation
to tune step-size parameters $(\tau_0, \kappa)$ in the stochastic gradient descent
component of the SVI algorithm. This cross-validation requires
multiple runs over the data and thus is not suited to the 
streaming setting. \figs{wiki_svi_step_size} and \figss{nature_svi_step_size}
demonstrate that the parameter choice does indeed affect 
algorithm performance. In these figures,
we keep $D$ at the true training data
size.

\cite{hoffman:2010:online} have observed that
the optimal
$(\tau_0, \kappa)$ may interact with minibatch size,
and we further observe that the optimal values may vary
with $D$ as well.
We also note that recent work has suggested a way to update $(\tau_0, \kappa)$
adaptively during an SVI run \cite{ranganath:2013:adaptive}.

\textbf{EP is not suited to LDA.}
Earlier attempts to apply EP to the LDA model in the non-streaming setting
have had mixed success, with \cite{buntine:2004:applying} in particular 
finding that EP performance can be poor for LDA and, moreover, that EP requires 
``unrealistic intermediate storage requirements.''  We found this to also
be true in the streaming setting.  We were not able to obtain competitive 
results with EP; based on an 8-thread implementation of SDA-Bayes with an 
EP primitive\footnote{We chose 8 threads since any fewer was too slow to 
get results and anything larger created too high of a memory demand on our system.},
after over 91 hours on Wikipedia (and $6.7 \times 10^{4}$ data points),
log predictive probability had stabilized at around $-7.95$
and, after over 97 hours on Nature (and $9.7 \times 10^{4}$ data points),
log predictive probability had stabilized at around $-8.02$.
Although SDA-Bayes with the EP primitive is not effective for LDA,
it remains to be seen whether this combination may be useful in other
domains where EP is known to be effective.

\section{Discussion} \label{sec:discussion}

We have introduced SDA-Bayes, a framework for streaming, distributed,
asynchronous computation of an approximate Bayesian posterior. Our 
framework makes streaming updates to the estimated posterior 
according to a user-specified approximation primitive.
We have demonstrated the usefulness of our framework,
with variational Bayes as the primitive, by fitting the 
latent Dirichlet allocation topic model to the Wikipedia and Nature
corpora.  We have demonstrated the advantages of our algorithm 
over stochastic variational inference and the sufficient statistics 
update algorithm, particularly with respect to the key issue of
obtaining approximations to posterior probabilities based on the
number of documents seen thus far, not posterior probabilities
for a fixed number of documents. 

\subsubsection*{Acknowledgments}

We thank M.~Hoffman, C.~Wang, and J.~Paisley
for discussions, code, and data and our reviewers for helpful comments. TB is supported by the Berkeley Fellowship, NB by a Hertz Foundation Fellowship, and ACW by the Chancellor's Fellowship at UC Berkeley. This research is supported in part by NSF CISE Expeditions award CCF-1139158, DARPA XData Award FA8750-12-2-0331, and AMPLab sponsor donations from Amazon Web Services, Google, SAP,  Blue Goji, Cisco, Clearstory Data, Cloudera, Ericsson, Facebook, General Electric, Hortonworks, Intel, Microsoft, NetApp, Oracle, Samsung, Splunk, VMware and Yahoo!. This material is based upon work supported in part by the Office of
Naval Research under contract/grant number N00014-11-1-0688.



\bibliographystyle{unsrt}
{\small \bibliography{obayes}}

\begin{thebibliography}{10}

\bibitem{niu:2011:hogwild}
F.~Niu, B.~Recht, C.~R{\'e}, and S.~J. Wright.
\newblock {H}ogwild!: A lock-free approach to parallelizing stochastic gradient
  descent.
\newblock In {\em Neural Information Processing Systems}, 2011.

\bibitem{kleiner:2012:big}
A.~Kleiner, A.~Talwalkar, P.~Sarkar, and M.~Jordan.
\newblock The big data bootstrap.
\newblock In {\em International Conference on Machine Learning}, 2012.

\bibitem{hoffman:2010:online}
M.~Hoffman, D.~M. Blei, and F.~Bach.
\newblock Online learning for latent {D}irichlet allocation.
\newblock In {\em Neural Information Processing Systems}, volume~23, pages
  856--864, 2010.

\bibitem{hoffman:2013:stochastic}
M.~Hoffman, D.~M. Blei, J.~Paisley, and C.~Wang.
\newblock Stochastic variational inference.
\newblock {\em Journal of Machine Learning Research}, 14:1303--1347.

\bibitem{wang:2011:online}
C.~Wang, J.~Paisley, and D.~M. Blei.
\newblock Online variational inference for the hierarchical {D}irichlet
  process.
\newblock In {\em Artificial Intelligence and Statistics}, 2011.

\bibitem{wainwright:2008:graphical}
M.~J. Wainwright and M.~I. Jordan.
\newblock Graphical models, exponential families, and variational inference.
\newblock {\em Foundations and Trends in Machine Learning}, 1(1-2):1--305,
  2008.

\bibitem{minka:2001:expectation}
T.~P. Minka.
\newblock Expectation propagation for approximate {B}ayesian inference.
\newblock In {\em Uncertainty in Artificial Intelligence}, pages 362--369.
  Morgan Kaufmann, 2001.

\bibitem{minka:2001:family}
T.~P. Minka.
\newblock {\em A family of algorithms for approximate {B}ayesian inference}.
\newblock PhD thesis, Massachusetts Institute of Technology, 2001.

\bibitem{opper:1998:bayesian}
M.~Opper.
\newblock A {B}ayesian approach to on-line learning.

\bibitem{canini:2009:online}
K.~R Canini, L.~Shi, and T.~L Griffiths.
\newblock Online inference of topics with latent {D}irichlet allocation.
\newblock In {\em Artificial Intelligence and Statistics}, volume~5, 2009.

\bibitem{honkela:2003:on-line}
A.~Honkela and H.~Valpola.
\newblock On-line variational {B}ayesian learning.
\newblock In {\em International Symposium on Independent Component Analysis and
  Blind Signal Separation}, pages 803--808, 2003.

\bibitem{luts:2012:real}
J.~Luts, T.~Broderick, and M.~P. Wand.
\newblock Real-time semiparametric regression.
\newblock {\em Journal of Computational and Graphical Statistics, to appear.
  Preprint arXiv:1209.3550.}

\bibitem{blei:2003:latent}
D.~M. Blei, A.~Y. Ng, and M.~I. Jordan.
\newblock Latent {D}irichlet allocation.
\newblock {\em Journal of Machine Learning Research}, 3:993--1022, 2003.

\bibitem{minka:2002:expectation}
T.~Minka and J.~Lafferty.
\newblock Expectation-propagation for the generative aspect model.
\newblock In {\em Uncertainty in Artificial Intelligence}, pages 352--359.
  Morgan Kaufmann, 2002.

\bibitem{teh:2006:collapsed}
Y.~Teh, D.~Newman, and M.~Welling.
\newblock A collapsed variational {B}ayesian inference algorithm for latent
  {D}irichlet allocation.
\newblock In {\em Neural Information Processing Systems}, 2006.

\bibitem{asuncion:2009:smoothing}
A.~Asuncion, M.~Welling, P.~Smyth, and Y.~Teh.
\newblock On smoothing and inference for topic models.
\newblock In {\em Uncertainty in Artificial Intelligence}, 2009.

\bibitem{hoffman:2010:online:code}
M.~Hoffman.
\newblock Online inference for {LDA} ({P}ython code) at \\
  \url{http://www.cs.princeton.edu/~blei/downloads/onlineldavb.tar}, 2010.

\bibitem{ranganath:2013:adaptive}
R.~Ranganath, C.~Wang, D.~M. Blei, and E.~P. Xing.
\newblock An adaptive learning rate for stochastic variational inference.
\newblock In {\em International Conference on Machine Learning}, 2013.

\bibitem{buntine:2004:applying}
W.~L. Buntine and A.~Jakulin.
\newblock Applying discrete {PCA} in data analysis.
\newblock In {\em Uncertainty in Artificial Intelligence}.

\bibitem{seeger:2005:expectation}
M.~Seeger.
\newblock Expectation propagation for exponential families.
\newblock Technical report, University of California at Berkeley, 2005.

\end{thebibliography}

\newpage

\appendix
\allowdisplaybreaks

\section{Variational Bayes} \label{app:vb}

\allowdisplaybreaks

\subsection{Batch VB} \label{app:batch_vb}

As described in the main text, the idea of VB is to find the distribution
$q_{D}$ that best approximates
the true posterior, $p_{D}$. More specifically,
the optimization problem of VB is defined as finding a $q_{D}$
to minimize
the KL divergence between the approximating distribution and the posterior:
$$
	\dkl{q_{D}}{p_{D}}
		:= \mbe_{q_{D}} \left[\log\left( q_{D} / p_{D}\right)\right]
$$
Typically $q_{D}$ takes a particular, constrained form, and 
finding the optimal $q_{D}$ amounts to finding the optimal parameters
for $q_{D}$. Moreover, the optimal parameters usually cannot be expressed
in closed form, so often a coordinate descent algorithm is used.

For the LDA model, we have $q_{D}$ in the form of
\eq{lda_vb_approx_q} and $p_{D}$ defined by \eq{lda_posterior}.
We wish to find the following variational parameters (i.e., parameters to $q_{D}$):
$\lambda$ (describing each topic), $\gamma$ (describing the
topic proportions in each document), and $\phi$ (describing the
assignment of each word in each document to a topic).

\subsubsection{Evidence lower bound}

Finding $q_{D}$ to minimize the KL divergence between $q_{D}$ and $p_{D}$ is equivalent
to finding $q_{D}$ to maximize the \emph{evidence lower bound} (ELBO),
\begin{align*}
	\elbo
		&:= \mbe_{q_{D}} \left[\log p(\Theta, x_{1:D})\right] - \mbe_{q_{D}} \left[\log q_{D}\right] \\
		&= \mbe_{q_{D}} \left[\log p_{D}\right] + p(x_{1:D}) - \mbe_{q_{D}} \left[\log q_{D}\right] \\
		&= -\dkl{q_{D}}{p_{D}} + p(x_{1:D}),
\end{align*}
since $p(x_{1:D})$ is constant in $q_{D}$.
The VB optimization problem is often phrased in terms of the ELBO
instead of the KL divergence. 

The ELBO for LDA can be written as follows, where the model
parameters are $\beta, \theta, z$ and the data is $w$; $\eta$ and $\alpha$
are fixed hyperparameters.
\begin{align*}
\elbo(\lambda, \gamma, \phi)
	&= \mbe_q\left[\log p(\beta,\theta,z,w \mid \eta, \alpha) \right] - \mbe_q\left[ \log q(\beta,\theta,z \mid \lambda, \gamma, \phi) \right] \\
	&= \sum_{k=1}^K \mbe_q\left[ \log \dir(\beta_k \mid \eta_k) \right]
		+ \sum_{d=1}^D \mbe_q\left[ \log \dir(\theta_d \mid \alpha) \right] \\
	&\qquad {} + \sum_{d=1}^D \sum_{n=1}^{N_d} \mbe_q\left[ \log \mult(z_{dn} \mid \theta_d) \right] 
		+ \sum_{d=1}^D \sum_{n=1}^{N_d} \mbe_q\left[ \log \mult(w_{dn} \mid \beta_{z_{dn}}) \right] \\
	&\qquad {} - \sum_{k=1}^K \mbe_q\left[ \log \dir(\beta_k \mid \lambda_k) \right]
		- \sum_{d=1}^D \mbe_q\left[ \log \dir(\theta_d \mid \gamma_d) \right] \\
	&\qquad - \sum_{d=1}^D \sum_{n=1}^{N_d} \mbe_q\left[ \log \mult(z_{dn} \mid \phi_{dw_{dn}}) \right].
\end{align*}
The expectations in $q$ in the previous equation can be evaluated as follows. The equations below
make use of the \emph{digamma function} $\psi$ and \emph{trigamma function}
$\psi_{1}$. Here,
\begin{align*}
	\psi(x) &= \frac{d}{dx} \log \Gamma(x) = \left[ \frac{d}{dx} \Gamma(x) \right] / \Gamma(x) \\
	\psi_{1}(x) &= \frac{d^{2}}{dx^{2}} \log \Gamma(x) = \frac{d}{dx} \psi(x).
\end{align*}
Then,
\begin{align*}
	\lefteqn{
	\mbe_q\left[ \log \dir(\beta_k \mid \eta_k) \right]
	} \\
		&\qquad = \log \Gamma\left(\sum_{v=1}^V \eta_{kv}\right)
			- \sum_{v=1}^V \log \Gamma(\eta_{kv})
			+ \sum_{v=1}^V (\eta_{kv}-1) \: \mbe_q[\log \beta_{kv}] \\
		&\qquad = \log \Gamma\left(\sum_{v=1}^V \eta_{kv}\right)
			- \sum_{v=1}^V \log \Gamma(\eta_{kv})
			+ \sum_{v=1}^V (\eta_{kv}-1) \left(\psi(\lambda_{kv})
			- \psi\Big(\sum_{u=1}^V \lambda_{ku}\Big) \right) \\
	\lefteqn{
	\mbe_q\left[ \log \dir(\theta_d \mid \alpha) \right]
	} \\
		&\qquad = \log \Gamma\left(\sum_{k=1}^K \alpha_k \right)
			- \sum_{k=1}^K \log \Gamma(\alpha_k)
			+ \sum_{k=1}^K (\alpha_k-1) \: \mbe_q[\log \theta_{dk}] \\
		&\qquad = \log \Gamma\left(\sum_{k=1}^K \alpha_k \right)
			- \sum_{k=1}^K \log \Gamma(\alpha_k)
			+ \sum_{k=1}^K (\alpha_k-1) \left(\psi(\gamma_{dk})
			- \psi\Big(\sum_{j=1}^K \gamma_{dj}\Big) \right) \\
	\lefteqn{
	\mbe_q\left[ \log \mult(z_{dn} \mid \theta_d) \right]
	} \\
		&\qquad = \sum_{k=1}^K \phi_{dw_{dn}k} \mbe_q[\log \theta_{dk}] \\
		&\qquad = \sum_{k=1}^K \phi_{dw_{dn}k} \left(\psi(\gamma_{dk})
			- \psi\Big(\sum_{j=1}^K \gamma_{dj}\Big) \right) \\
	\lefteqn{
	\mbe_q\left[ \log \mult(w_{dn} \mid \beta_{z_{dn}}) \right]
	} \\
		&\qquad = \sum_{v=1}^V \mbo\{w_{dn}=v\} \: \mbe_q[\log \beta_{z_{dn},v}] \\
		&\qquad = \sum_{v=1}^V \mbo\{w_{dn}=v\} \: \sum_{k=1}^K \phi_{dw_{dn}k} \mbe_q[\log \beta_{kv}] \\
		&\qquad = \sum_{v=1}^V \sum_{k=1}^K \mbo\{w_{dn}=v\} \:\phi_{dw_{dn}k} \left(\psi(\lambda_{kv})
			- \psi\Big(\sum_{u=1}^V \lambda_{ku}\Big) \right) \\
	\lefteqn{
	\mbe_q\left[ \log \dir(\beta_k \mid \lambda_k) \right]
	} \\
		&\qquad = \log \Gamma\left(\sum_{v=1}^V \lambda_{kv} \right)
			- \sum_{v=1}^V \log \Gamma(\lambda_{kv})
			+ \sum_{v=1}^V (\lambda_{kv}-1) \: \mbe_q[\log \beta_{kv}] \\
		&\qquad = \log \Gamma\left(\sum_{v=1}^V \lambda_{kv} \right)
			- \sum_{v=1}^V \log \Gamma(\lambda_{kv})
			+ \sum_{v=1}^V (\lambda_{kv}-1) \left(\psi(\lambda_{kv})
			- \psi\Big(\sum_{u=1}^V \lambda_{ku}\Big) \right) \\
	\lefteqn{
	\mbe_q\left[ \log \dir(\theta_d \mid \gamma_d) \right]
	} \\
		&\qquad = \log \Gamma\left(\sum_{k=1}^K \gamma_{dk} \right)
			- \sum_{k=1}^K \log \Gamma(\gamma_{dk})
			+ \sum_{k=1}^K (\gamma_{dk}-1) \: \mbe_q[\log \theta_{dk}] \\
		&\qquad = \log \Gamma\left(\sum_{k=1}^K \gamma_{dk} \right)
			- \sum_{k=1}^K \log \Gamma(\gamma_{dk})
			+ \sum_{k=1}^K (\gamma_{dk}-1) \left(\psi(\gamma_{dk})
			- \psi\Big(\sum_{j=1}^K \gamma_{dj}\Big) \right) \\
	\lefteqn{
	\mbe_q\left[ \log \mult(z_{dn} \mid \phi_{dn}) \right]
	} \\
		&\qquad = \sum_{k=1}^K \phi_{dw_{dn}k} \log \phi_{dw_{dn}k}.
\end{align*}

\subsubsection{Coordinate ascent}

We maximize the ELBO via coordinate ascent in each dimension of the 
variational parameters: $\lambda$, $\gamma$, and $\phi$.

\paragraph{Variational parameter $\lambda$.}
Choose a topic index $k$.
Fix $\gamma$, $\phi$, and each $\lambda_j$ for $j \ne k$. Then we can write the ELBO's functional
dependence on $\lambda_k$ as follows, where ``$\const$'' is a constant in $\lambda_k$.
\begin{align*}
	\elbo(\lambda_k)
	&= \sum_{v=1}^V (\eta_{kv}-1)
			\left(\psi(\lambda_{kv}) - \psi\Big(\sum_{u=1}^V \lambda_{ku}\Big) \right) \\
	&\qquad {} + \sum_{d=1}^D \sum_{n=1}^{N_d} \sum_{v=1}^V \mbo\{w_{dn}=v\} \: \phi_{dw_{dn}k}
			\left(\psi(\lambda_{kv}) - \psi\Big(\sum_{u=1}^V \lambda_{ku}\Big) \right) \\
	&\qquad {} - \log \Gamma\left(\sum_{v=1}^V \lambda_{kv} \right)
		+ \sum_{v=1}^V \log \Gamma(\lambda_{kv}) \\
	&\qquad {} - \sum_{v=1}^V (\lambda_{kv}-1)
			\left(\psi(\lambda_{kv}) - \psi\Big(\sum_{u=1}^V \lambda_{ku}\Big) \right)
			+ \const \\
	&= \sum_{v=1}^V \left( \eta_{kv} - \lambda_{kv} + \sum_{d=1}^D \sum_{n=1}^{N_d} \mbo\{w_{dn}=v\} \: \phi_{dw_{dn}k} \right)
			\left(\psi(\lambda_{kv}) - \psi\Big(\sum_{u=1}^V \lambda_{ku}\Big) \right) \\
	&\qquad {} - \log \Gamma\left(\sum_{v=1}^V \lambda_{kv} \right)
		+ \sum_{v=1}^V \log \Gamma(\lambda_{kv})
		+ \const
\end{align*}
The partial derivative of $\elbo(\lambda_k)$ with respect to one of the dimensions of $\lambda_k$, say $\lambda_{kv}$, is
\begin{align*}
	\lefteqn{
	\frac{\partial}{\partial \lambda_{kv}} \elbo(\lambda_k)
	} \\
	&\qquad = - \left(\psi(\lambda_{kv}) - \psi\Big(\sum_{u=1}^V \lambda_{ku}\Big) \right) \\
	&\qquad \qquad {} + \left( \eta_{kv} - \lambda_{kv} + \sum_{d=1}^D \sum_{n=1}^{N_d} \mbo\{w_{dn}=v\} \: \phi_{dw_{dn}k} \right) \left(\psi_1(\lambda_{kv})
		- \psi_1\Big(\sum_{u=1}^V \lambda_{ku}\Big) \right) \\
	&\qquad \qquad {} - \sum_{t: t \neq v} \left( \eta_{kt} - \lambda_{kt} + \sum_{d=1}^D \sum_{n=1}^{N_d} \mbo\{w_{dn}=t\} \: \phi_{dw_{dn}k} \right) \: \psi_1\Big(\sum_{u=1}^V \lambda_{ku}\Big)
		- \psi\Big(\sum_{u=1}^V \lambda_{ku}\Big) + \psi(\lambda_{kv}) \\
	&\qquad = \psi_1(\lambda_{kv}) \: \left( \eta_{kv} - \lambda_{kv} + \sum_{d=1}^D \sum_{n=1}^{N_d} \mbo\{w_{dn}=v\} \: \phi_{dw_{dn}k} \right) \\
	&\qquad \qquad {} - \psi\Big(\sum_{u=1}^V \lambda_{ku}\Big) \: \sum_{u=1}^V \left( \eta_{ku} - \lambda_{ku} + \sum_{d=1}^D \sum_{n=1}^{N_d} \mbo\{w_{dn}=u\} \: \phi_{dw_{dn}k} \right).
\end{align*}
From the last line of the previous equation, we see that one can set the gradient of $\elbo(\lambda_k)$ to zero by setting
$$
	\lambda_{kv} \leftarrow \eta_{kv} + \sum_{d=1}^D \sum_{n=1}^{N_d} \mbo\{w_{dn}=v\} \: \phi_{dw_{dn}k} \quad \text{ for } v = 1,\dots,V.
$$
Equivalently, if $n_{dv}$ is the number of occurrences (tokens) of word type $v$ in document $d$,
then the update may be written
$$
	\lambda_{kv} \leftarrow \eta_{kv} + \sum_{d=1}^D n_{dv} \; \phi_{dvk} \quad \text{ for } v = 1,\dots,V.
$$

\paragraph{Variational parameter $\gamma$.}
Now choose a document $d$. Fix $\lambda$, $\phi$, and $\gamma_{c}$ for $c \ne d$. Then we
can express the functional dependence of the ELBO on $\gamma_d$ as follows.
\begin{align*}
	\elbo(\gamma_d)
		&= \sum_{k=1}^K (\alpha_k-1) \left(\psi(\gamma_{dk}) - \psi\Big(\sum_{j=1}^K \gamma_{dj}\Big) \right)
			+ \sum_{n=1}^{N_d} \sum_{k=1}^K \phi_{dw_{dn}k} \left(\psi(\gamma_{dk}) - \psi\Big(\sum_{j=1}^K \gamma_{dj}\Big) \right) \\
		&\qquad {} - \log \Gamma\left(\sum_{k=1}^K \gamma_{dk} \right)
			+ \sum_{k=1}^K \log \Gamma(\gamma_{dk})
			- \sum_{k=1}^K (\gamma_{dk}-1) \left(\psi(\gamma_{dk})
			- \psi\Big(\sum_{j=1}^K \gamma_{dj}\Big) \right) \\
		&\qquad {} + \const \\
		&= \sum_{k=1}^K \left( \alpha_k - \gamma_{dk} + \sum_{n=1}^{N_d} \phi_{dw_{dn}k} \right) 						\left(\psi(\gamma_{dk}) - \psi\Big(\sum_{j=1}^K \gamma_{dj}\Big) \right) \\
		&\qquad {} - \log \Gamma\left(\sum_{k=1}^K \gamma_{dk} \right)
			+ \sum_{k=1}^K \log \Gamma(\gamma_{dk})
			+ \const
\end{align*}
The partial derivative of $\elbo(\gamma_d)$ with respect to one of the dimensions of $\gamma_d$, say $\gamma_{dk}$, is
\begin{align*}
	\lefteqn{
	\frac{\partial}{\partial \gamma_{dk}} \elbo(\gamma_d)
	} \\
		&= -\left(\psi(\gamma_{dk})
			- \psi\Big(\sum_{j=1}^K \gamma_{dj}\Big) \right)
			+ \left( \alpha_k - \gamma_{dk} + \sum_{n=1}^{N_d} \phi_{dw_{dn}k} \right)
				\left(\psi_1(\gamma_{dk}) - \psi_1\Big(\sum_{j=1}^K \gamma_{dj}\Big) \right) \\
		&\qquad {} - \sum_{i: i \neq k} \left( \alpha_i - \gamma_{di} + \sum_{n=1}^{N_d} \phi_{dw_{dn}i} \right) \:
				\psi_1\Big(\sum_{j=1}^K \gamma_{dj}\Big)
			- \psi\Big(\sum_{j=1}^K \gamma_{dj}\Big)
			+ \psi(\gamma_{dk}) \\
		&= \psi_1(\gamma_{dk}) \left( \alpha_k - \gamma_{dk} + \sum_{n=1}^{N_d} \phi_{dw_{dn}k} \right)
			- \psi_1\Big(\sum_{j=1}^K \gamma_{dj}\Big)
				\sum_{j=1}^K \left( \alpha_j - \gamma_{dj} + \sum_{n=1}^{N_d} \phi_{dw_{dn}j} \right).
\end{align*}
As for the $\lambda$ case above, one obvious way to achieve a gradient of $\elbo(\gamma_d)$ equal to zero is to set
$$
	\gamma_{dk} \leftarrow \alpha_k + \sum_{n=1}^{N_d} \phi_{dw_{dn}k} \quad \text{ for } k = 1,\dots,K.
$$
Equivalently,
$$
	\gamma_{dk} \leftarrow \alpha_k + \sum_{v=1}^{V} n_{dv} \; \phi_{dvk} \quad \text{ for } k = 1,\dots,K.
$$

\paragraph{Variational parameter $\phi$.}
Finally, consider fixing $\lambda$, $\gamma$, and $\phi_{cu}$ for $(c, u) \ne (d, v)$.
In this case, the dependence of the ELBO on $\phi_{dv}$ can be written as follows.
\begin{align*}
	\lefteqn{
	\elbo(\phi_{dv})
	} \\
	&= \sum_{k=1}^K n_{dv} \; \phi_{dvk} \left(\psi(\gamma_{dk}) - \psi\Big(\sum_{j=1}^K \gamma_{dj}\Big) \right) \\
	&\qquad {} + \sum_{k=1}^K n_{dv} \:\phi_{dvk}
			\left(\psi(\lambda_{kv}) - \psi\Big(\sum_{u=1}^V \lambda_{ku}\Big) \right)
		- \sum_{k=1}^K n_{dv} \; \phi_{dvk} \log \phi_{dvk}
		+ \const \\
	&= \sum_{k=1}^K n_{dv} \; \phi_{dvk} \left( - \log \phi_{dvk} + \psi(\gamma_{dk}) - \psi\Big(\sum_{j=1}^K \gamma_{dj}\Big) + \psi(\lambda_{kv}) - \psi\Big(\sum_{u=1}^V \lambda_{ku}\Big)\right) \\
	&\qquad {} + \const
\end{align*}
The partial derivative of $\elbo(\phi_{dv})$ with respect to one of the dimensions of $\phi_{dv}$, say $\phi_{dvk}$, is
\begin{align*}
	\lefteqn{
	\frac{\partial}{\partial \phi_{dvk}} \elbo(\phi_{dv})
	} \\
	&= n_{dv}
		\left(- \log \phi_{dvk}
			+ \psi(\gamma_{dk})
			- \psi\Big(\sum_{j=1}^K \gamma_{dj}\Big)
			+ \psi(\lambda_{kv})
			- \psi\Big(\sum_{u=1}^V \lambda_{ku}\Big)
			- 1
		\right).
\end{align*}
Using the method of Lagrange multipliers to incorporate the constraint that $\sum_{k=1}^K \phi_{dvk} = 1$,
we wish to find $\rho$ and $\phi_{dvk}$ such that
\begin{equation}
	\label{eq:grad_wrt_phi_set_0}
	0 = \frac{\partial}{\partial \phi_{dvk}} \left[ \elbo(\phi_{dv}) - \rho \left(\sum_{k=1}^K \phi_{dvk} - 1 \right) \right].
\end{equation}
Setting
\begin{align*}
	\phi_{dvk}
		\propto_k \exp
			\left(
				\psi(\gamma_{dk}) - \psi\Big(\sum_{j=1}^K \gamma_{dj}\Big)
				+ \psi(\lambda_{kv}) - \psi\Big(\sum_{u=1}^V \lambda_{ku}\Big) 
			\right)
\end{align*}
achieves the desired outcome in \eq{grad_wrt_phi_set_0}. Here, $\propto_k$ indicates that the proportionality is across $k$. The optimal choice of $\rho$ is expressed via this proportionality.
The above assignment may also be written as
\begin{align*}
	\phi_{dvk}
		\propto_k \exp
			\left( 
				\mbe_{q}[ \log \theta_{dk} ]
				+ \mbe_{q}[ \log \beta_{kv} ]
			\right)
\end{align*}

The coordinate-ascent algorithm iteratively updates the parameters $\lambda$, $\gamma$, and $\phi$. In practice, we usually iterate the updates for the ``local'' parameters $\phi$ and $\gamma$ until they converge, then update the ``global'' parameter $\lambda$, and repeat. The resulting batch variational Bayes algorithm is presented in \alg{batchvb_lda}.

\subsection{SDA-Bayes VB} \label{app:sda_vb_lda}

For a fixed hyperparameter $\alpha$, we can think of $\batchvb$ as an algorithm
that takes input in the form of a prior on topic parameters $\beta$ and a minibatch of
documents. In particular, let $C_{b}$ be the $b$th minibatch of documents;
for documents with indices in $\mathcal{D}_{b}$, these documents can be
summarized by the word counts $(n_{d})_{d \in \mathcal{D}_{b}}$. Then, in the 
notation of \eq{seq_bayes_update}, we have $\Theta = \beta$, $\mathcal{A} = \batchvb$, 
and
\begin{align*}
	q_{0}(\beta) &= \prod_{k=1}^{K} \dir(\beta_k | \eta_k).
\end{align*}
In general, the $b$th posterior takes the same form and
therefore can be summarized by its parameters
$\lambda^{(b)}$:
\begin{align*}
	q_{b}(\beta) &= \prod_{k=1}^{K} \dir(\beta_k | \lambda^{(b)}_k).
\end{align*}

In this case, if we set the prior parameters to $\lambda^{(0)}_k := \eta_k$, \eq{seq_bayes_update} becomes the following algorithm.

\begin{algorithm}[H]
	\KwIn{Hyperparameter $\eta$}
	Initialize $\lambda^{(0)} \leftarrow \eta$ \\
	\ForEach{Minibatch $C_{b}$ of documents}{
		$\lambda^{(b)} \leftarrow \batchvb\Big( C_{b}, \lambda^{(b-1)} \Big)$ \\
		$q_b(\beta) = \prod_{k=1}^{K} \dir(\beta_k | \lambda^{(b)}_k)$
	}
\caption{\label{alg:sda_stream_vb_lda} Streaming VB for LDA}
\end{algorithm}	

Next, we apply the asynchronous, distributed updates described in the
``Asynchronous Bayesian updating'' portion of \mysec{bayes_update} to 
the batch VB primitive and LDA model. In this case, $\lambda^{(\post)}$ is the posterior
parameter estimate maintained at the master, and each worker updates this value 
after a local computation. The posterior after seeing a collection of minibatches
is $q(\beta) = \prod_{k=1}^{K} \dir(\beta_k | \lambda^{(\post)}_k)$.

\begin{algorithm}[H]
	\KwIn{Hyperparameter $\eta$}
	Initialize $\lambda^{(\post)} \leftarrow \eta$ \\
	\ForEach{Minibatch $C_{b}$ of documents, at a worker}{
		Copy master value locally: $\lambda^{(local)} \leftarrow \lambda^{(\post)}$
		$\lambda \leftarrow \batchvb\Big( C_{b}, \lambda^{(\local)} \Big)$ \\
		$\Delta \lambda \leftarrow \lambda - \lambda^{(\local)}$ \\
		Update the master value synchronously: $\lambda^{(\post)} \leftarrow \lambda^{(\post)} + \Delta \lambda$
	}
\caption{\label{alg:sda_vb_lda} SDA-Bayes with VB primitive for LDA}
\end{algorithm}	

\section{Expectation Propagation} \label{app:ep}

\subsection{Batch EP} \label{app:batch_ep}

Our batch expectation propagation (EP) algorithm for LDA
learns a posterior for both the document-specific topic mixing proportions
$(\theta_{d})_{d=1}^{D}$
and the topic distributions over words $(\beta_{k})_{k=1}^{K}$.
By contrast, the algorithm in \cite{minka:2002:expectation}
learns only the former and so is not appropriate to the 
model in \mysec{lda}.

For consistency, we also follow \cite{minka:2002:expectation} in 
making a distinction between token and type word updates, where
a token refers to a particular word instance and a type refers to
all words with the same vocabulary value. Let $C = (w_d)_{d=1}^{D}$ denote the set of documents
that we observe, and for each word $v$ in the vocabulary, let $n_{dv}$ denote the number of times $v$ appears in document $d$.

\paragraph{Collapsed posterior.}
We begin by collapsing (i.e., integrating out) the word assignments $z$ in the posterior~\eqref{eq:lda_posterior} of LDA. We can express the collapsed
posterior as
\begin{equation*}
	p(\beta,\theta \mid C, \eta,\alpha)
		\propto \left[ \prod_{k=1}^K \dir_{V}(\beta_k \mid \eta_k) \right]
			\cdot \prod_{d=1}^D \left[ \dir_{K}(\theta_d \mid \alpha)
				\cdot \prod_{v=1}^V \left(
				\sum_{k=1}^K \theta_{dk} \: \beta_{kv} \right)^{n_{dv}} \right].
\end{equation*}
For each document-word pair $(d,v)$, consider approximating the term $\sum_{k=1}^K \theta_{dk} \beta_{kv}$ above by
\begin{equation*}
	\left[ \prod_{k=1}^K \dir_{V}(\beta_k \mid \bloc_{kdv}+\mathbf{1}_{V}) \right] \cdot \dir_{K}(\theta_d \mid \tloc_{dv}+\mathbf{1}_{K}),
\end{equation*}
where $\bloc_{kdv} \in \R^V$, $\tloc_{dv} \in \R^K$, and $\mathbf{1}_{M}$ is a vector of all ones of length $M$. This proposal serves as inspiration for taking the approximating variational distribution for $p(\beta,\theta \mid C,\eta,\alpha)$
to be of the form
\begin{equation}
	\label{eq:approx_ep}
	q(\beta,\theta \mid \lambda, \gamma)
		:= \left[ \prod_{k=1}^K q(\beta_k \mid \lambda_k) \right]
			\cdot \prod_{d=1}^D q(\theta_d \mid \gamma_d),
\end{equation}
where $q(\beta_k \mid \lambda_k) = \dir(\beta_k \mid \lambda_k)$ and $q(\theta_d \mid \gamma_d) = \dir(\theta_d \mid \gamma_d)$, with the parameters
\begin{equation}
	\label{eq:ep_param_decomposition}
	\lambda_k = \eta_k + \sum_{d=1}^D \sum_{v=1}^V n_{dv} \bloc_{kdv}, \qquad
	\gamma_d = \alpha + \sum_{v=1}^V n_{dv} \tloc_{dv},
\end{equation}
and the constraints $\lambda_k \in \Rp^V$ and $\gamma_d \in \Rp^K$ for each $k$ and $d$.
We assume this form in the remainder of the analysis and write $q(\beta, \theta \mid \bloc, \tloc)$ for $q(\beta, \theta \mid \lambda, \gamma)$, where $\bloc = (\bloc_{kdv})$, $\tloc = (\tloc_{dv})$.

\textbf{Optimization problem.}
We seek to find the optimal parameters $(\bloc, \tloc)$ by minimizing the (reverse) KL divergence:
\begin{equation*}
	\min_{\bloc, \tloc} \; \dkl{p(\beta,\theta \mid C,\eta,\alpha)}{q(\beta,\theta \mid \bloc, \tloc)}.
\end{equation*}
This joint minimization problem is not tractable, and the idea of EP is to proceed iteratively by fixing most of the factors in \eq{approx_ep} and minimizing the KL divergence over the parameters related to a single word.

More formally, suppose we already have a set of parameters $(\bloc, \tloc)$. Consider a document
$d$ and word $v$ that occurs in document $d$ (i.e., $n_{dv} \geq 1$). We start by removing
the component of $q$ related to
$(d,v)$ in \eq{approx_ep}. Following~\cite{minka:2001:expectation}, we subtract out the effect of one occurrence of word $v$ in document $d$, but at the end of this process we update the distribution on the type level. In doing so, we use the following shorthand for the remaining global parameters:
\begin{align*}
	\bpdv_k &= \lambda_k - \bloc_{kdv} = \eta_k + (n_{dv}-1) \bloc_{kdv} + \sum_{(d',v'): (d',v') \neq (d,v)} n_{d'v'} \bloc_{kd'v'} \\ 
	\tpdv_d &= \gamma_{d} - \tloc_{dv} = \alpha + (n_{dv}-1) \tloc_{dv} + \sum_{v': v' \neq v} n_{dv'} \tloc_{dv'} 
	.
\end{align*}
We replace this removed part of $q$ by the term $\sum_{k=1}^K \theta_{dk} \beta_{kv}$, which corresponds to the contribution of one occurrence of word $v$ in document $d$ to the true posterior $p$. Call the resulting normalized distribution $\tilde{q}_{dv}$, so $\tilde q_{dv}(\beta, \theta \mid \bpdv, \gamma_{\setminus d}, \tpdv_d)$ satisfies
\begin{align*}
	\propto \left[ \prod_{k=1}^K \dir(\beta_k \mid \bpdv_k) \right]
		\cdot \left[\prod_{d'\neq d} \dir(\theta_{d'} \mid \gamma_{d'} ) \right]
		\cdot \dir(\theta_d \mid \tpdv_d) \cdot \sum_{k=1}^K \theta_{dk} \: \beta_{kv}.
\end{align*}
We obtain an improved estimate of the posterior $q$ by updating the parameters from $(\lambda,\gamma)$ to $(\hat \lambda, \hat \gamma)$, where
\begin{equation}
	\label{eq:ep_update_principle}
	(\hat \lambda, \hat \gamma)
		= \arg\min_{\lambda',\gamma'} \; \dkl{
				\tilde q_{dv}(\beta,\theta \mid \bpdv, \gamma_{\setminus d}, \tpdv_d)
			}{
				q(\beta,\theta \mid \lambda', \gamma')
			}.
\end{equation}

\textbf{Solution to the optimization problem.}
First, note that for $d': d' \ne d$, we have $\hat \gamma_{d'} = \gamma_{d'}$.

Now consider the index $d$ chosen on this iteration. Since $\beta$ and $\theta$ are Dirichlet-distributed
under $q$, the minimization problem in \eq{ep_update_principle} reduces to solving the moment-matching equations \cite{minka:2001:expectation,seeger:2005:expectation}
\begin{align*}
	\mbe_{\tilde q_{dv}}[\log \beta_{ku}]
		&= \mbe_{\hat \lambda_k}[\log \beta_{ku}]
			\qquad \text{ for } 1 \leq k \leq K, \; 1 \leq u \leq V, \\
	\mbe_{\tilde q_{dv}}[\log \theta_{dk}]
		&= \mbe_{\hat \gamma_d}[\log \theta_{dk}] \qquad\; \text{ for } 1 \leq k \leq K
	.
\end{align*}
These can be solved via Newton's method
though \cite{minka:2001:expectation} recommends
solving exactly for the first and ``average second'' moments
of $\beta_{ku}$ and $\theta_{dk}$, respectively, instead.
We choose the latter approach for consistency with
\cite{minka:2001:expectation}; our own experiments also
suggested taking the approach of \cite{minka:2001:expectation} was
faster than Newton's method with no noticeable performance loss.
The resulting moment updates are
\begin{align}
	\label{eq:ep_moment_matching_beta}
	\hat \lambda_{ku}
		&= \frac{
				\sum_{y=1}^{V} \left( \mbe_{\tilde{q}_{dv}}[\beta^{2}_{ky}]
					- \mbe_{\tilde{q}_{dv}}[\beta_{ky}] \right)
			}{
				\sum_{y=1}^{V} \left( \mbe_{\tilde{q}_{dv}}[\beta_{ky}]^{2}
					- \mbe_{\tilde{q}_{dv}}[\beta^{2}_{ky}] \right)
			}
			\cdot \mbe_{\tilde{q}_{dv}}[\beta_{ku}] \\
	\label{eq:ep_moment_matching_theta}
	\hat \gamma_{dk}
		&= \frac{
				\sum_{j=1}^{K} \left( \mbe_{\tilde{q}_{dv}}[\theta^{2}_{dj}]
					- \mbe_{\tilde{q}_{d,n}}[\theta_{dj}] \right)
			}{
				\sum_{j=1}^{K} \left( \mbe_{\tilde{q}_{dv}}[\theta_{dj}]^{2}
					- \mbe_{\tilde{q}_{dv}}[\theta^{2}_{dj}] \right)
			}
			\cdot \mbe_{\tilde{q}_{dv}}[\theta_{dk}]
	.
\end{align}
We then set $(\bloc_{kdv})_{k=1}^{K}$ and $\tloc_{dv}$ such that the new global parameters $(\lambda_k)_{k=1}^{K}$ and $\gamma_d$ are equal to the optimal parameters $(\hat \lambda_k)_{k=1}^{K}$ and $\hat \gamma_d$. The resulting algorithm is presented below (Alg.~\ref{alg:ep_lda}).

\begin{algorithm}[H]
	\KwIn{Data $C = (w_d)_{d=1}^{D}$; hyperparameters $\eta, \alpha$}
	\KwOut{$\lambda$}
	Initialize $\forall (k,d,v)$, $\bloc_{kdv} \leftarrow 0$ and $\tloc_{dv} \leftarrow 0$ \\
	\While{$(\bloc, \tloc)$ not converged}{
		\ForEach{$(d,v)$ with $n_{dv} \geq 1$}{
			\vspace{1mm}
			\tcc{Variational distribution without the word token $(d,v)$}
			$\forall k, \: \bpdv_k \leftarrow \eta_k + (n_{dv}-1) \bloc_{kdv} + \sum_{(d',v') \neq (d,v)} n_{d'v'} \bloc_{kd'v'}$ \\
			$\tpdv_d \leftarrow \alpha + (n_{dv}-1) \tloc_{dv} + \sum_{v' \neq v} n_{dv'} \tloc_{dv'}$ \\
			If any of $\bpdv_{ku}$ or $\tpdv_{dk}$ are non-positive, skip updating this $(d,v)$ \hfill $(\dagger)$\\
			\vspace{1mm}
			\tcc{Variational parameters from moment-matching}
			$\forall (k,u)$, compute $\hat{\lambda}_{ku}$ from \eq{ep_moment_matching_beta} \\
			$\forall k$, compute $\hat{\gamma}_{dk}$ from \eq{ep_moment_matching_theta} \\
			\vspace{1mm}
			\tcc{Type-level updates to parameter values}
			$\forall k, \: \bloc_{kdv} \leftarrow n_{dv}^{-1} \left(\hat{\lambda}_{k} - \bpdv_{k}\right) + \left(1 - n_{dv}^{-1} \right) \bloc_{kdv}$ \\
			$\tloc_{dv} \leftarrow n_{dv}^{-1} \left( \hat{\gamma}_{d} - \tpdv_{d} \right) + \left(1 - n_{dv}^{-1} \right) \tloc_{dv}$ \\
			Other $\bloc, \tloc$ remain unchanged
		}
	}
	\tcc{Global variational parameters}
	$\forall k, \: \lambda_k \leftarrow \eta_k + \sum_{d=1}^D \sum_{v=1}^V n_{dv} \bloc_{kdv}$
	\caption{\label{alg:ep_lda} EP for LDA}
\end{algorithm}

The results in the main text (\mysec{evaluation}) are reported for \alg{ep_lda}. We also tried a slightly modified EP algorithm that makes token-level updates to parameter values, rather than type-level updates. This modified version iterates through each word \emph{placeholder} in document $d$; that is, through pairs $(d,n)$ rather than pairs $(d,v)$ corresponding to word \emph{values}. Since there are always at least as many $(d,n)$ pairs as $(d,v)$ pairs with $n_{dv} \ge 1$ (and usually many more of the former), the modified algorithm requires many more iterations. In practice, we find better experimental performance for the modified EP algorithm in terms of log predictive probability as a function of number of data points in the training set seen so far: e.g., leveling off at about $-7.96$ for Nature vs.\ $-8.02$. However, the modified algorithm is also much slower, and still returns much worse results than SDA-Bayes or SVI, so we do not report these results in the main text.\footnote{Here and in the main text we run EP with $\eta = 1$. We also tried EP with $\eta = 0.01$, but the positivity check for $\bpdv_{ku}$ and $\tpdv_{dk}$ on line~$(\dagger)$ in Algorithm~\ref{alg:ep_lda} always failed and as a result none of the parameters were updated.}

\subsection{SDA-Bayes EP} \label{app:sda_ep_lda}

Putting a batch EP algorithm for LDA into the SDA-Bayes framework
is almost identical to putting a batch VB algorithm for LDA into the SDA-Bayes
framework. This similarity is to be expected since SDA-Bayes works out of the 
box with a batch approximation algorithm in the correct form.

For a fixed hyperparameter $\alpha$, we can think of $\batchep$ as an algorithm
(just like $\batchvb$)
that takes input in the form of a prior on topic parameters $\beta$ and a minibatch of
documents. The same
setup and notation from \app{sda_vb_lda} applies. In this case,
\eq{seq_bayes_update} becomes the following algorithm.

\begin{algorithm}[H]
	\KwIn{Hyperparameter $\eta$}
	Initialize $\lambda^{(0)} \leftarrow \eta$ \\
	\ForEach{Minibatch $C_{b}$ of documents}{
		$\lambda^{(b)} \leftarrow \batchep\Big( C_{b}, \lambda^{(b-1)} \Big)$ \\
		$q_b(\beta) = \prod_{k=1}^{K} \dir(\beta_k | \lambda^{(b)}_k)$
	}
\caption{\label{alg:sda_stream_ep_lda} Streaming EP for LDA}
\end{algorithm}	

This algorithm is exactly the same as \alg{sda_stream_vb_lda} but with a
batch EP primitive instead of a batch VB primitive.

Next, we apply the asynchronous, distributed updates described in the ``Asynchronous Bayesian updating'' portion of \mysec{bayes_update} to 
the batch EP primitive and LDA model. Again, the setup and notation from
\app{sda_vb_lda} applies, and we find the following algorithm.

\begin{algorithm}[H]
	\KwIn{Hyperparameter $\eta$}
	Initialize $\lambda^{(\post)} \leftarrow \eta$ \\
	\ForEach{Minibatch $C_{b}$ of documents, at a worker}{
		Copy master value locally: $\lambda^{(local)} \leftarrow \lambda^{(\post)}$
		$\lambda \leftarrow \batchep\Big( C_{b}, \lambda^{(\local)} \Big)$ \\
		$\Delta \lambda \leftarrow \lambda - \lambda^{(\local)}$ \\
		Update the master value synchronously: $\lambda^{(\post)} \leftarrow \lambda^{(\post)} + \Delta \lambda$
	}
\caption{\label{alg:sda_ep_lda} SDA-Bayes with EP primitive for LDA}
\end{algorithm}	

Indeed, the recipe outlined here applies more generally to other primitives besides EP and VB.

\end{document}